\documentclass[10pt,twocolumn,letterpaper]{article}

\usepackage{wacv}              

\usepackage[T1]{fontenc}
\usepackage{graphicx}
\usepackage{amsmath}
\usepackage{amssymb}
\usepackage[T1]{fontenc}
\usepackage[dvipsnames]{xcolor}
\usepackage{array}
\usepackage{booktabs}
\usepackage{makecell}
\usepackage{multirow}
\usepackage{siunitx}

\usepackage{algorithm}
\usepackage{algorithmicx}
\usepackage{algpseudocode}
\usepackage[misc]{ifsym}

\newcolumntype{C}[1]{>{\centering\let\newline\\\arraybackslash\hspace{0pt}}m{#1}}

\usepackage[pagebackref,breaklinks,colorlinks]{hyperref}

\def\theoden{\href{https://github.com/MECLabTUDA/TheODen}{https://github.com/MECLabTUDA/TheODen}}

\algdef{SE}[SUBALG]{Indent}{EndIndent}{}{\algorithmicend\ }%
\algtext*{Indent}
\algtext*{EndIndent}

\definecolor{Ventricle}{rgb}{0.7372549019607844, 0.3137254901960784, 0.5647058823529412}
\definecolor{Midline}{rgb}{0.34509803921568627, 0.3137254901960784, 0.5529411764705883}

\usepackage[capitalize]{cleveref}
\crefname{section}{Sec.}{Secs.}
\crefname{algorithm}{Alg.}{Algos.}
\Crefname{section}{Section}{Sections}
\Crefname{table}{Table}{Tables}
\crefname{table}{Tab.}{Tabs.}

\def\wacvPaperID{1722} 
\def\confName{WACV}
\def\confYear{2025}

\begin{document}

\title{Federated Voxel Scene Graph for Intracranial Hemorrhage}



\author{
Antoine P. Sanner$^{1,2}$ \textsuperscript{(\Letter)}, Jonathan Stieber$^1$, Nils F. Grauhan$^2$, Suam Kim$^2$,\\ Marc A. Brockmann$^2$, Ahmed E. Othman$^2$, Anirban Mukhopadhyay$^1$\\
$^1$ Department of Computer Science, Technical University of Darmstadt, Germany\\
$^2$ Department of Neuroradiology, University Medical Center Mainz, Germany\\
{\tt\small antoine.sanner@gris.tu-darmstadt.de}
}

\index{Antoine Sanner}
\index{Jonathan Stieber}
\index{Nils Grauhan}
\index{Suam Kim}
\index{Marc Brockmann}
\index{Ahmed Othman}
\index{Anirban Mukhopadhyay}

\maketitle

\begin{abstract}

Intracranial Hemorrhage is a potentially lethal condition whose manifestation is vastly diverse and shifts across clinical centers worldwide.
Deep-learning-based solutions are starting to model complex relations between brain structures, but still struggle to generalize. 
While gathering more diverse data is the most natural approach, privacy regulations often limit the sharing of medical data.
We propose the first application of Federated Scene Graph Generation.
We show that our models can leverage the increased training data diversity. For Scene Graph Generation, they can recall up to $20$\% more clinically relevant relations across datasets compared to models trained on a single centralized dataset. 
Learning structured data representation in a federated setting can open the way to the development of new methods that can leverage this finer information to regularize across clients more effectively.

\end{abstract}

\section{Introduction}
\label{sec:intro}

\begin{figure*}
  \centering
  \begin{subfigure}[t]{0.24\linewidth}
    \centering
    \includegraphics[width=.9\linewidth]{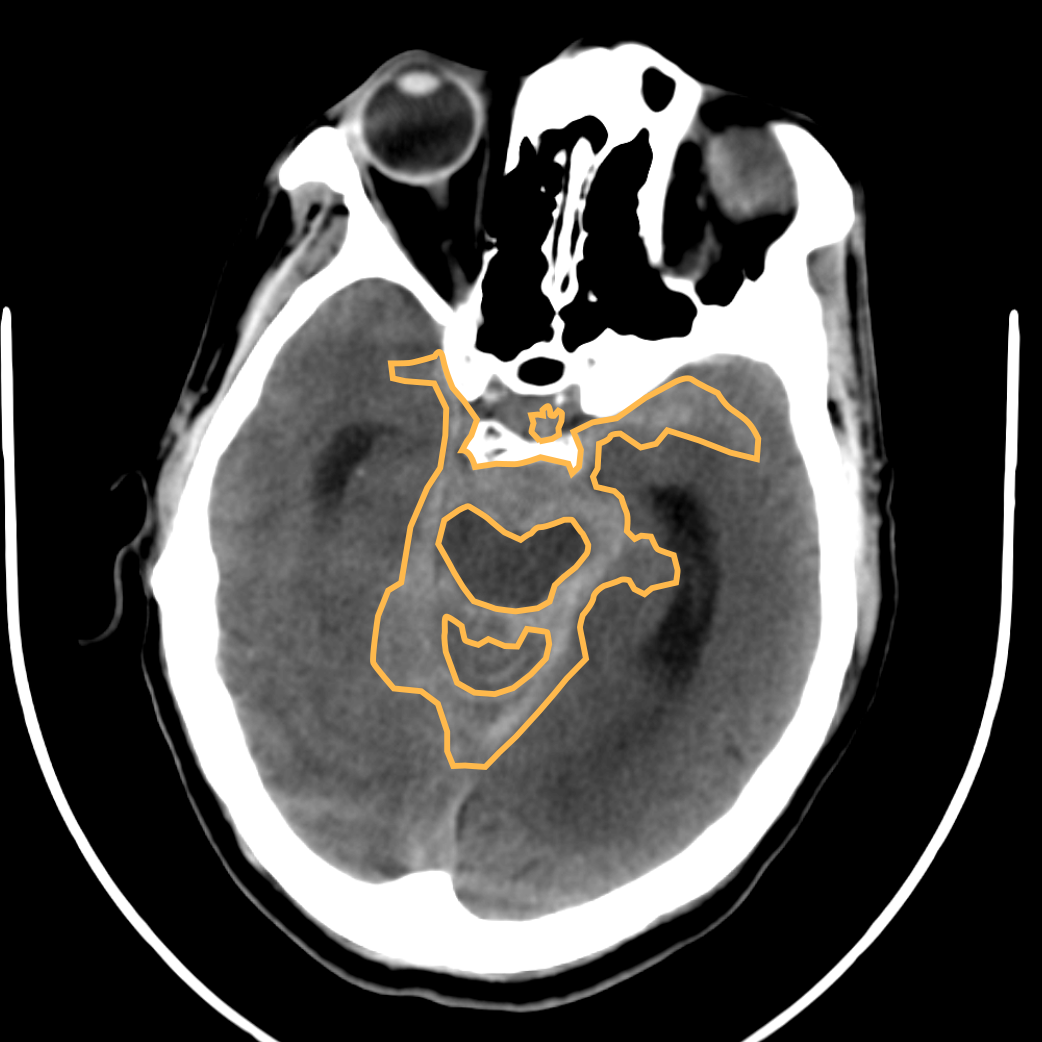}
    \caption{Spontaneous subarachnoid hemorrhage}
  \end{subfigure}
  \hfill
  \begin{subfigure}[t]{0.24\linewidth}
    \centering
    \includegraphics[width=.9\linewidth]{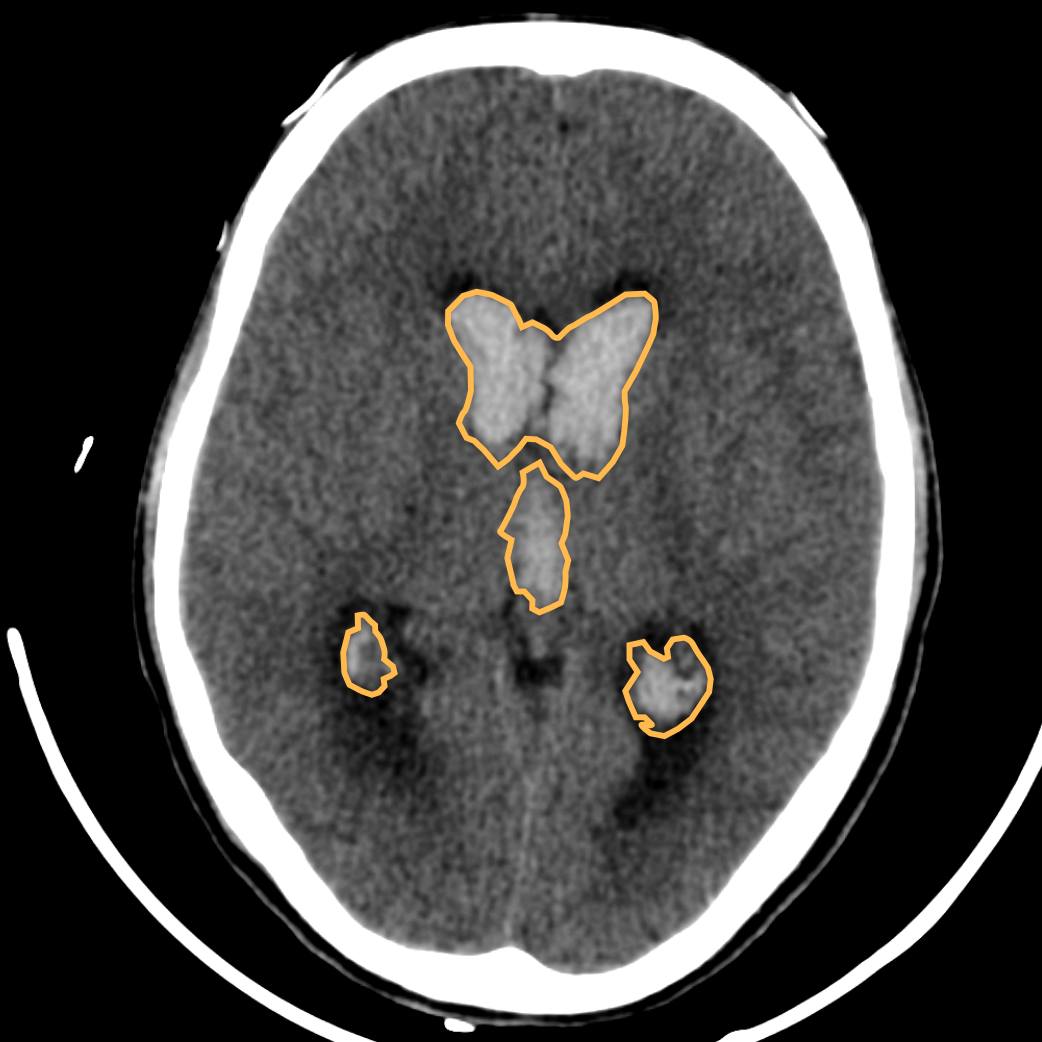}
    \caption{Intraventricular hemorrhage}
  \end{subfigure}
  \hfill
  \begin{subfigure}[t]{0.24\linewidth}
    \centering
    \includegraphics[width=.9\linewidth]{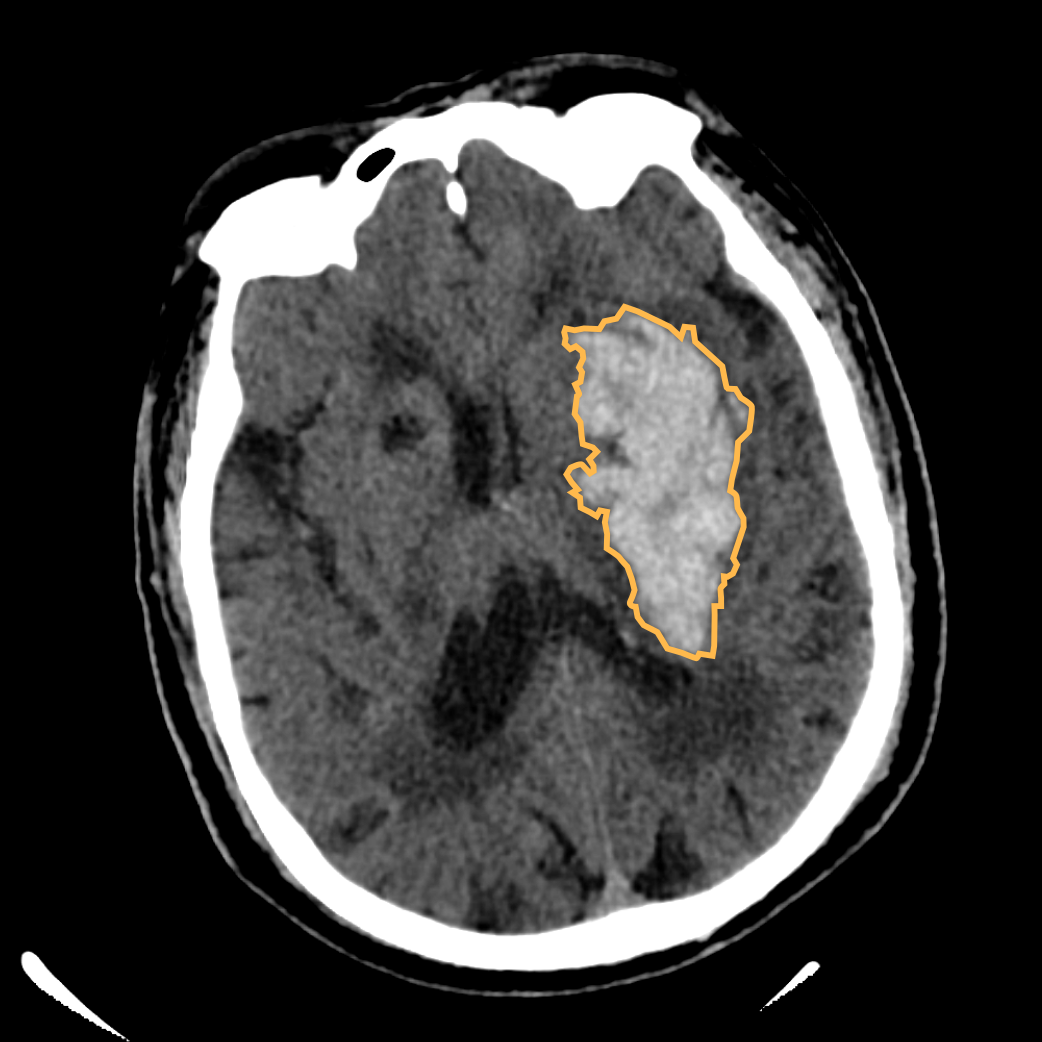}
    \caption{Intraparenchymal hemorrhage}
  \end{subfigure}
  \hfill
  \begin{subfigure}[t]{0.24\linewidth}
    \centering
    \includegraphics[width=.9\linewidth]{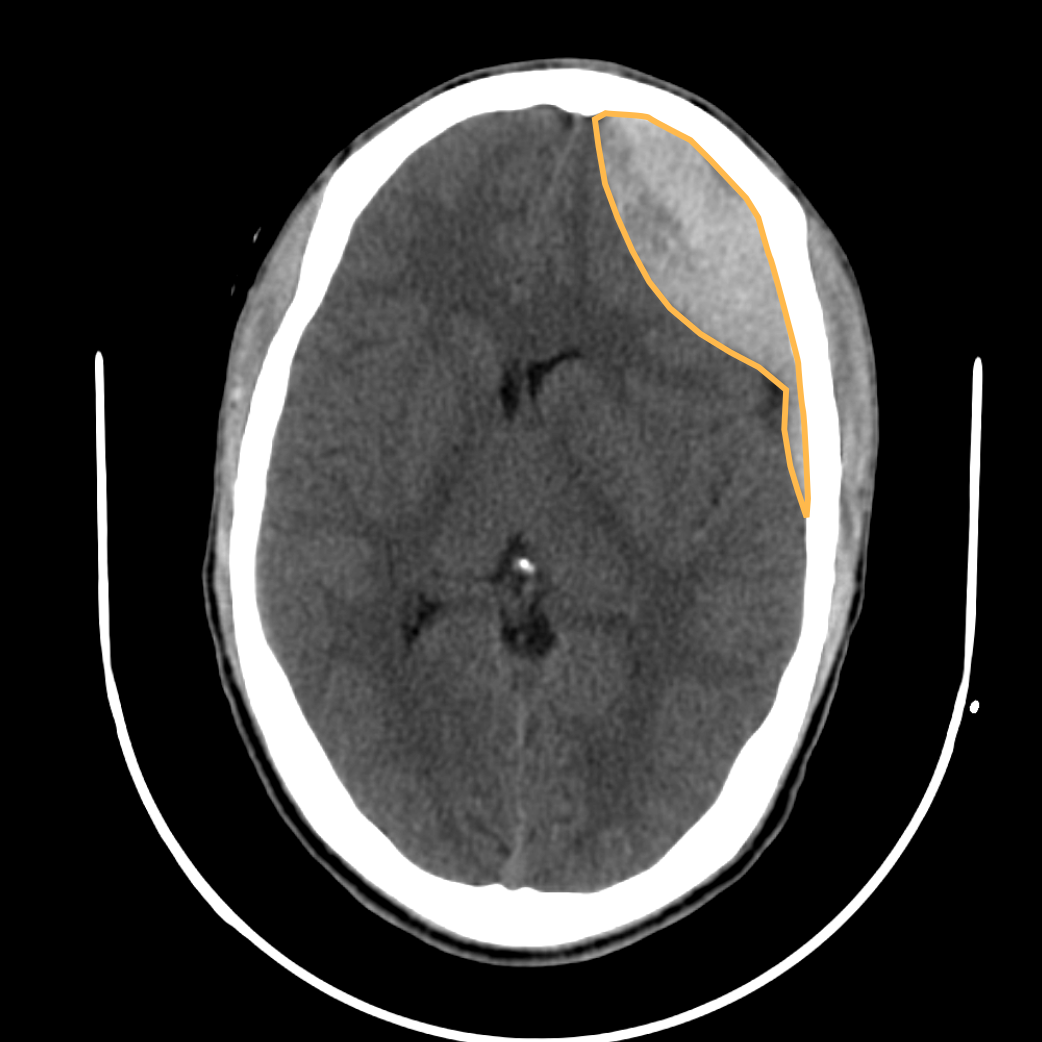}
    \caption{Epidural hemorrhage in a trauma patient.}
  \end{subfigure}
  \caption{
  Examples of the diversity in manifestation of ICH. The outline of the bleeding is shown in \textcolor{Dandelion}{yellow}. Hemorrhages such as in (a) may require a surgical intervention to repair any ruptured blood vessel or the placement of a drainage to relieve pressure. Similarly, involvement of the ventricular system as in (b) can cause occlusive hydrocephalus and will also require a drainage for the accumulating cerebrospinal fluid. Intraparenchymal (c) and epidural (d) hemorrhages, while dissimilar in appearance, can both cause midline shifts (c and d). Such a shift is associated with increased intracranial pressure and may require surgery.
  }
  \label{fig:ich_diversity}
\end{figure*}

Intracranial Hemorrhage (ICH) is a potentially lethal condition, which requires swift detection and treatment to improve the patients' odds of survival \cite{Greenberg2022-wo,Hemphill2015-ke}. However, the term ICH captures a variety of situations. For instance, hypertension can cause the spontaneous rupture of a blood vessel and lead to a subarachnoidal hemorrhage under the brain. In contrast, trauma patients will more often have a subdural or epidural hemorrhage along the skull. 
The difference for such cases are visualized in \cref{fig:ich_diversity}. Treatment decisions remain clinically challenging as they need to be 1) patient-centered despite the diversity of manifestations of ICH and 2) swift as the patient outcome worsens shortly after ICH onset \cite{Al-Shahi_Salman2018-he}. The clinical routine involves the acquisition of head CTs for diagnosis
. We can employ Deep Learning (DL) on such images to support clinicians in their decisions and  improve the treatment of ICH patients.

Clinical centers worldwide see local shifts in disease manifestation, which makes it problematic for purely supervised representation learning to perform to its full potential. Especially with patient data privacy, available data is often scarce. Federated Learning (FedL) of visual representation has gained traction in recent years \cite{Guendouzi2023}, as it enables learning to address diversity issues without sharing the private data. This is especially relevant for medical data, since many hospitals own data and have capacities to gather annotations, but these data are usually patient data that need to be protected. By using FedL, one can leverage the heterogeneity of the available data to improve the models' generalizability while preserving the privacy of the patients. However, pure visual representation learning even using FedL only offers a superficial understanding of the clinical case, especially compared to the structured approach clinicians often use.

\begin{figure*}
  \centering
  \includegraphics[width=\linewidth]{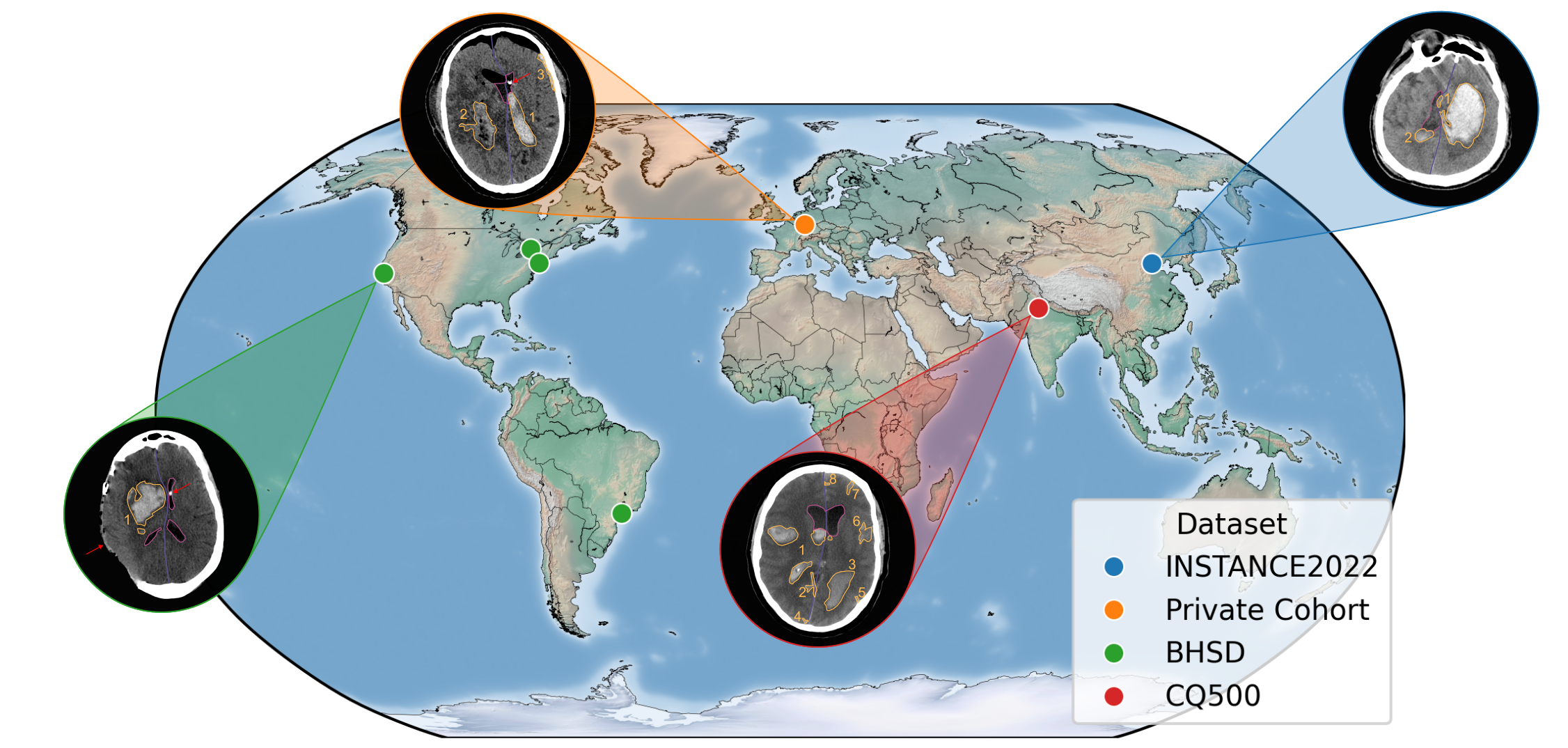}
  \caption{
  Overview of the origin and diversity of the four datasets used for this study: INSTANCE2022, BHSD, CQ500, and a private cohort from Germany.
  We show the outline of \textcolor{Dandelion}{ICH}, the \textcolor{Ventricle}{ventricle system}, and \textcolor{Midline}{midline}. \textcolor{Dandelion}{Bleeding 1} from INSTANCE2022, CQ500, and the private cohort all involve the \textcolor{Ventricle}{ventricle system}, which often serves as a buffer for other brain structures. The \textcolor{Ventricle}{ventricle system} can compress to absorb external pressure, or conversely it can fill with blood with possible expansion. Such changes are often accompanied by a \textcolor{Midline}{midline shift}, as in the samples of the INSTANCE2022, BHSD and private cohort datasets.
  Additionally, some images show the results of a previous surgical operation such as the presence of a ventricular drainage (appearing as a white dot within the slice) or even a craniectomy, see the \textcolor{Red}{red} arrows. \cref{sec:exp} offers detailed statistics over these cohorts. 
  }
  \label{fig:fl_setup}
\end{figure*}

The majority of existing DL work focuses on the detection or segmentation of ICH \cite{Cho2019-az,Emon2024,Heit2021-jo,Kuo2019-ix,li2023stateoftheart,sanner2024detectionintracranialhemorrhagetrauma,spiegler2024weaklysupervisedintracranialhemorrhage,Wang2021-vc,wu2023bhsd3dmulticlassbrain,Xu2024}. These models are trained using centralized learning on individual datasets, and often fail to generalize well to other data distributions. While segmentation is useful for computing the volume of the hemorrhage, it is ill-suited for the detection of individual bleeding \cite{sanner2024detectionintracranialhemorrhagetrauma}. Even detecting ICH accurately is not enough from a clinical perspective, as no clinical complication caused by the bleeding are modeled. The involvement of the ventricular system through hemorrhage expansion or the bleeding-induced shift of midline can occur and are both strong predictors of poor patient outcome \cite{Kuo2019-ix,Garton2016-zg,Xu2023-ay}. The clinical utility of DL lies in analyzing the structure of the \textbf{clinical cerebral scene} using a specialized representation.
Recently, Voxel Scene Graph Generation (V-SGG) \cite{sanner2024voxelscenegraphintracranial} has shown promising results in modeling the clinical cerebral scene through a structured representation incorporating both ICH localization and the relations between ICH and adjacent brain structures.
Likewise to other studies using Centralized Learning, the models detected $24$\% fewer relations when evaluated for Scene Graph Generation on an external cohort with a tangible data shift.

We introduce Federated Voxel Scene Graph Generation. Motivated by \textit{Neural Motifs} \cite{DBLP:journals/corr/abs-1711-06640} and \textit{Iterative Message Passing} \cite{xu2017scene}, we propose the \textbf{Fed-MOTIF} and \textbf{Fed-IMP} methods, which learn a common relation distribution across clients in a federated setup and to minimize the bias towards client-local distributions.
We validate our methods on four datasets originating from all over the world. \cref{fig:fl_setup} gives an overview of the data origins, as well as how anatomically dissimilar two ICH cases can be. Nevertheless, clinical decisions still depend on the same set of complex relations, independently of the precise ICH manifestation.
Our models trained with FedL can recall up to $20$\% more clinically relevant relations compared to models trained on a single centralized dataset for Scene Graph Generation.
With this work, we pioneer Federated Voxel Scene Graph\footnote{Code available at \url{https://github.com/MECLabTUDA/VoxelSceneGraph}}, which generalizes across four datasets sourced worldwide and offer improved ICH detection for each bleeding type.


\section{Related Work}

This section first presents existing work on ICH detection, whether through segmentation, image classification or pure object detection. We then introduce aggregation methods used for Federated Learning.

\subsection{Intracranial Hemorrhage Detection}

The earlier works on DL applied to ICH detection \cite{Cho2019-az,Heit2021-jo,Kuo2019-ix,Wang2021-vc,Xu2024-vr} focus on predicting the presence or absence of bleeding. Such methods offer the most elemental information about a patient's case, but do not offer any insights in ICH volume or localization. 

Subsequent works \cite{Emon2024,physionet,li2023stateoftheart,spiegler2024weaklysupervisedintracranialhemorrhage,wu2023bhsd3dmulticlassbrain,Xu2024} tackle ICH segmentation, particularly as ICH volume is a strong predictor of patient outcome \cite{Pinho2019-fq}. While these methods offer an improvement over previously used coarse volume estimation methods such as ABC/2 \cite{Kleinman2011-bv}, segmentation is ill-suited to offer precise ICH localization \cite{DBLP:journals/corr/abs-1811-08661,sanner2024detectionintracranialhemorrhagetrauma}. A connected component analysis can compute bounding boxes out of a predicted segmentation masks, but such a method is prone to failures. Indeed, an under-segmented bleeding may appear as multiple components and result in multiple detections. Similarly, bleedings with a satellite sign \cite{Deng2019} are clinically perceived as a single bleeding, although it is composed of multiple blood masses. Recent studies showed that precise ICH localization is possible through bounding box prediction \cite{sanner2024detectionintracranialhemorrhagetrauma,spiegler2024weaklysupervisedintracranialhemorrhage}. Nevertheless, the presented work fails to model further than the presence or localization of ICH.

ICH is an anomaly, that takes valuable space within the finite skull volume. As such, it often interacts with neighboring brain structures. The involvement of the ventricular system through hemorrhage expansion or the bleeding-induced shift of midline can cause severe clinical complications \cite{Kuo2019-ix,Garton2016-zg,Xu2023-ay}.  The clinical utility of DL lies in modeling such relations between individual bleedings and neighboring brain structures. Voxel Scene Graph Generation \cite{sanner2024voxelscenegraphintracranial} has shown promising results in modeling the clinical cerebral scene. Nevertheless, the models in this study and in all others presented before are trained on single datasets. The few studies, which investigate model robustness, repeatedly show failures to generalize to other cohorts containing distribution shifts. \cite{chilamkurthy2018developmentvalidationdeeplearning,sanner2024detectionintracranialhemorrhagetrauma,sanner2024voxelscenegraphintracranial,Wang2021-vc}.

\subsection{Federated Learning}

Federated Learning changes the training paradigm to learn one or multiple models collaboratively using multiple entities, referred to as clients. Most importantly, the training data is not centralized, and each client only has access to their own data. 
A central server aggregates the local models trained on each client into a global model. FedL methods mainly differ in how the aggregation operation is performed. \textit{FedAvg} \cite{mcmahan2023communicationefficientlearningdeepnetworks} offers a straightforward approach by computing the average of all client models. Other methods, such as \textit{FedSGD} and \textit{FedAdam} \cite{reddi2021adaptivefederatedoptimization} leverage classical model optimizers at a server-level to guide the aggregation.
Most FedL applications focus on visual representation learning \cite{Lu2024}, but none on structure Scene Graph learning.

\section{Method}

In this section, we first describe SGG for Intracranial Hemorrhage. Then, we introduce our method for Federated Object Detection and Federated Scene Graph Generation. \cref{alg:meth_client,alg:meth_serv} give insights into the federated learning process for V-SGG
. The inference process only consists of a forward pass through the entire global model.

\subsection{Scene Graph for Intracranial Hemorrhage}
\label{sec:vsgg}

For any SGG application, a Knowledge Graph needs to be first defined using domain knowledge to guide the annotation process. Sanner et al. \cite{sanner2024voxelscenegraphintracranial} lay the groundwork of V-SGG for ICH. They select three classes of objects: 1) \emph{bleeding}, 2) v\emph{entricle system}, and 3) \emph{midline}, with each patient having exactly one ventricle system and one midline. Additionally, three possible clinical complications are modeled through three bleeding-induced relation classes: 1) \emph{midline shifts}, 2) \emph{blood flow to the ventricle system}, and 3) swelling induced \emph{asymmetry of the ventricle system}. With this set of relation classes, bleeding instances without any relation to other brain structures can swiftly be ruled as of lesser clinical importance.

\begin{algorithm}
\caption{
Training for Fed-SGG, \textbf{client-side}. We define the functions executed by each client to perform training steps using their corresponding \textbf{local data}.
}
\label{alg:meth_client}
\begin{algorithmic}
\State \textbf{Input:} \begin{itemize}
    \item Clients indexed by $c$ with annotated local training data $\mathcal{D}^c_{obj}$, and a subset $\mathcal{D}^c_{rel} \subseteq \mathcal{D}^c_{obj}$ containing only images with relations
    \item Loss functions for object detection and relation prediction $\ell_{loc}$, $\ell_{box}$, $\ell_{seg}$, $\ell_{rel}$
    \item $K$ max object detections per image
    \item Learning rates $\eta_{obj}$, $\eta_{rel}$
    \item Number of steps per round $E_{obj}$, $E_{rel}$
\end{itemize}

\item

\State \textbf{ClientUpdateObjDetec}($c$, $w_0^c$):
\Indent
\For{each step $e = 0...E_{obj}-1$}
    \State Sample batch $b^c$ from $\mathcal{D}^c_{obj}$
    \State $w_{e+1}^c \gets w_e^c - \eta_{obj} \nabla(\ell_{loc}(w_e^c,b^c) + \ell_{box}(w_e^c,b^c) + \ell_{seg}(w_e^c,b^c))$
\EndFor
\State Return $w_{E_{obj}}^c$ to the server
\EndIndent

\item

\State \textbf{ClientStatsUpdate}($c$, $w$):
\Indent
\State Compute confusion matrix $f^c$ over relations in $\mathcal{D}^c_{rel}$
\State Set the weights of the frequency bias in $w$ to $f^c$
\State Return $w$ to the server
\EndIndent

\item

\State \textbf{ClientUpdateRelPred}($c$, $w_0^c$):
\Indent
\For{each step $e = 0...E_{rel}-1$}
    \State Sample batch $b^c$ from $\mathcal{D}^c_{rel}$
    \State Detect up to $K$ objects $O_i=\{o_{i,0},...,o_{i,K-1}\}$ using $w_e^c$ for each image $i$ in $b^c$
    \State Select subject-object pairs for each image $P_i=\{s_{i,j}, o_{i,k}\}_{s_{i,j}, o_{i,k} \in O_i^2\}}$
    \State $w_{e+1}^c \gets w_e^c - \eta_{rel} \nabla\ell_{rel}(w_e^c,\{P_i\}_{i \in b^c})$
\EndFor
\State Return $w_{E_{rel}}^c$ to the server
\EndIndent

\end{algorithmic}
\end{algorithm}

\begin{algorithm}
\caption{
Training for Fed-SGG, \textbf{server-side}. We define the \textbf{main training pipeline} that is executed and scheduled by the server.
}
\label{alg:meth_serv}
\begin{algorithmic}
\State \textbf{Input:} \begin{itemize}
    \item $N$ clients indexed by $c$
    \item A weight aggregation algorithm \textbf{agg}, e.g. \textit{FedAvg}
    \item Number of training rounds $T_{obj}$, $T_{rel}$
\end{itemize}

\State{\bf Server executes:}
\Indent

\State Initialize global model $w_0$

\item
\State \# Perform object detection training
\For{each round $t = 0...T_{obj}-1$}
    \For{each client $c = 1...N$ \textbf{in parallel}}
        \State $w_{t+1}^c \gets$ \textbf{ClientUpdateObjDetec}$(c, w_t)$
    \EndFor
    \State $w_{t+1} \gets$ \textbf{agg}$(\{w_{t+1}^c\}_{c\in[1,N]})$
\EndFor
\State Freeze object detector of $w_{T_{obj}}$

\item
\State \# Initialize relation statistics
\For{each client $c = 1...N$ \textbf{in parallel}}
    \State $w_{T_{obj}}^c \gets$ \textbf{ClientStatsUpdate}$(c, w_{T_{obj}})$
\EndFor
\State $w_{T_{obj}} \gets$ mean$(\{w_{T_{obj}}^c\}_{c\in[1,N]})$

\item
\State \# Perform relation prediction training
\For{each round $t = T_{obj}...T_{obj}+T_{rel}-1$}
    \For{each client $c = 1...N$ \textbf{in parallel}}
        \State $w_{t+1}^c \gets$ \textbf{ClientUpdateRelPred}$(c, w_t)$
    \EndFor
    \State $w_{t+1} \gets$ \textbf{agg}$(\{w_{t+1}^c\}_{c\in[1,N]})$
\EndFor

\EndIndent

\end{algorithmic}
\end{algorithm}

\subsection{Federated Object Detection}

In a first step, the localization of relevant objects is predicted. On one side, we have the ventricle system and the midline, which are singletons present in every volume. Additionally, the midline is long, and tall, but very thin. This makes anchor-based detection challenging. On the other side, bleedings can differ in shape, position and scale and as such require a multiscale approach. The 3D Retina-UNet \cite{DBLP:journals/corr/abs-1711-06640} can solve both issues by design. This architecture can simultaneously predict bounding boxes at different scales using its Feature Pyramid Network and offer finer structure localization through semantic segmentation. The latter can be used to robustly localize the ventricle system and midline. Additionally, we combine this architecture with the novel \textit{VC-IoU} loss for bounding-box regression, which showed improved model performance over other losses for ICH detection \cite{sanner2024detectionintracranialhemorrhagetrauma}.

As the Retina-UNet is anchor-based, a set of anchors needs to be defined, which needs to fit all datasets. Sanner et al. \cite{sanner2024detectionintracranialhemorrhagetrauma} introduce a bleeding-adapted family of anchors. We employ this one for our studies after verifying that the number of matches between anchors and ground truth bounding boxes remains consistent across datasets. As this evaluation can be done by each client locally, it does not infringe any data privacy rules.

During the first step of the federated training process, only the object detector is trained. The server initializes a global model and propagates it to all clients. The anchors used are propagated as well as non-trainable layers in the model state. The training then occurs over $T_{obj}$ training rounds with $E_{obj}$ steps each, where a first loss is minimized. It has three components: an anchor classification term $\ell_{loc}$ for object localization, a box regression term $\ell_{box}$ to refine anchor shapes, and a segmentation quality term $\ell_{seg}$.

\subsection{Federated Voxel Scene Graph Generation}

Scene Graph Generation is the task of predicting relations between objects in an image, where a relation takes the form of a \textit{Subject-Predicate-Object} triplet. 
V-SGG methods build an enriched object, and relation representation by allowing information to flow through the entire scene graph before a final classification step. \textit{Neural Motifs} (\textit{MOTIF}) \cite{DBLP:journals/corr/abs-1711-06640} and \textit{Iterative Message Passing} (\textit{IMP}) \cite{xu2017scene} have been recently introduced to model relations in voxel data. The former uses bidirectional Long Short-Term Memory Networks \cite{Hochreiter1997} and iterates over detected objects and relations. In contrast, the latter combines Gated Recurrent Units \cite{cho2014propertiesneuralmachinetranslation} with message passing iteratively. Both methods use a relation frequency bias over the training distribution to refine all predictions.

To carry over to Federated Voxel Scene Graph Generation, we design federated variants (\textbf{Fed-MOTIF} and \textbf{Fed-IMP}) to learn an estimate of relation classes distribution across clients without the sharing of any data.
This frequency global bias allows the models to be more robust and less skewed towards a local distribution. 

In the second and last stage of the federated training, we consider that the object detector is fully trained. As such, the server freezes the weights related to object detection. Then each client computes statistics over the local relation distribution and the server aggregates all into a global statistics. Then the training is resumed for $T_{rel}$ rounds with $E_{rel}$ steps each. Only the $\ell_{rel}$ loss is minimized, which only models the classification of relations between object pairs (binary cross-entropy loss).
Contrary to SGG applications on natural images, we do not allow for the re-classification of predicted objects. \cref{alg:meth_client,alg:meth_serv} summarize the training pipeline.

\begin{table*}
    \centering
    \begin{tabular}{l cccc cc}
    \toprule
    &\multicolumn{4}{c}{\textbf{Object Detection}} & \multicolumn{2}{c}{\textbf{Relation Prediction}}\\
    \cmidrule(lr){2-5}\cmidrule(lr){6-7}
    & Ventricle & Midline & \multicolumn{2}{c}{Bleeding} & \multicolumn{2}{c}{Upper Bounds}\\
    \cmidrule(lr){2-2}\cmidrule(lr){3-3}\cmidrule(lr){4-5}\cmidrule(lr){6-7}
    Method & AR$_{30}\uparrow$ & AR$_{30}\uparrow$ & AR$_{30}\uparrow$ & AP$_{30}\uparrow$ & R@8$\uparrow$ & mR@8$\uparrow$\\
    \midrule
Avg. seen & 96.7±2.0 & 94.5±2.6 & 52.8±9.4 & 43.7±11.5 & 76.9±6.6 & 79.2±5.9\\
Avg. unseen & 82.0±26.5 & 78.7±22.6 & 44.9±15.5 & 35.0±16.3 & 61.4±23.9 & 64.8±24.1\\
\midrule
\textit{FedAvg} & 97.1±2.1 & \bf 95.5±1.8 & \bf 60.5±9.9 & \bf 51.5±12.2 & \bf 84.6±6.4 & \bf 88.2±3.9\\
\textit{FedSGD} & \bf 97.6±1.9 & 92.8±4.2 & 59.4±10.9 & 50.5±12.1 & 82.9±7.2 & 86.8±5.5\\
\midrule
All seen & 97.1±2.4 & 93.4±3.2 & 57.5±11.4 & 48.7±12.7 & 79.9±8.5 & 84.3±6.7\\
\bottomrule
\end{tabular}
\caption{
Left: Results for object detection for Centralized and Federated training.
Right: Upper bounds for relation prediction given the objects that are detected.
\textit{Avg. seen} and \textit{Avg. unseen} respectively refer to the results on in-distribution and out-of-distribution testing of models trained on only one dataset.
\textit{FedAvg}, and \textit{FedSGD} are trained on all datasets using Federated Learning.
\textit{All seen} refers to oracle models trained on all datasets in a centralized setup.
All configurations are run 5 times using random seeds.
}
\label{res:obj_detec}
\end{table*}

\section{Experimental Setup}
\label{sec:exp}

In this section, we present the datasets used for our study, as well as our annotation process.
We then move on to describe our evaluation setup and metrics. 
Details on data preprocessing, model training, and implementation are available in the Supplementary Material.

\subsection{Datasets}

To simulate the federated training of V-SGG models for ICH, we utilize three publicly available datasets for the segmentation or detection of ICH: INSTANCE2022 (INST) \cite{li2023stateoftheart}, BHSD \cite{wu2023bhsd3dmulticlassbrain}, and CQ500 \cite{chilamkurthy2018developmentvalidationdeeplearning}.
We decide to exclude the PhysioNet dataset \cite{physionet} as 1) there is strong distribution shift of the appearance of the ventricle system and midline compared to the other 3 datasets due to the patient cohort being much younger, and 2) this dataset only contains 26 cases with ICH.
The selected datasets stem from clinical centers from all over the world. We then perform some additional data curation to exclude images with either no visible ICH or a significant presence of acquisition artifacts, e.g. 1) movement artifacts causing streaks within the image or the misalignment of consecutive slices, and 2) metal artifacts. As a result, we select 120 images from INST, 100 from BHSD, and 157 from CQ500 for this study. Additionally, we build a private cohort (PC) with 67 patients from Germany and diagnosed with ICH. The Supplementary Material shows how the four datasets diverge both in terms of bleeding representation and clinical relations.

\subsection{Annotation Process}

First, a medical student produces a label map based on the segmentation computed during the pre-processing stage using \textit{3D Slicer} \cite{Fedorov2012-qh}.
More precisely, their task is to 1) fix any mistake in the segmentation masks, 2) segment any bleeding that was missed and 3) to produce a preliminary split of individual bleedings. A senior neuroradiologist then controls both the quality of ICH segmentation and its split. We modify the publicly available ICH masks released for the INST and BHSD datasets to annotate a few missed bleedings, but also to annotate visible blood even with low contrast. The primary purpose is to reduce the annotation bias across datasets and to focus rather on visual shifts across datasets. The final version of the ICH masks used for the INST and BHSD datasets have respectively a Dice score of 83.2±12.0\% and 81.3±19.1\% when compared to their original version.
In a last stage, two senior neuroradiologists use an in-house tool to annotate relations and the ICH type of each bleeding. This step also serves as an additional quality control checkpoint to ensure that no bleeding has been missed.
We will make the annotation tools and annotated data available in a separate publication.

\begin{table*}
    \centering
    \begin{tabular}{ll ccc ccc}
    \toprule
    &&\multicolumn{3}{c}{\textbf{Predicate Classification}} & \multicolumn{3}{c}{\textbf{Scene Graph Generation}}\\
    \cmidrule(lr){3-5}\cmidrule(lr){6-8}
    Method & Model & R@8$\uparrow$ & mR@8$\uparrow$ & mAP@8$\uparrow$ & R@8$\uparrow$ & mR@8$\uparrow$ & mAP@8$\uparrow$\\
    \midrule
Avg. seen & \textit{MOTIF} & 64.7±10.8 & 64.7±11.4 & 57.8±15.5 & 44.6±10.4 & 47.2±12.2 & 34.5±13.6\\
Avg. seen & \textit{IMP} & 60.0±7.5 & 59.3±9.3 & 53.7±14.9 & 48.3±11.0 & 48.0±12.5 & 18.1±8.9\\
Avg. unseen & \textit{MOTIF} & 54.5±17.8 & 54.7±18.8 & 54.3±16.1 & 32.3±17.5 & 34.1±19.7 & 24.3±16.5\\
Avg. unseen & \textit{IMP} & 52.2±15.2 & 52.0±17.8 & 47.4±17.0 & 35.5±18.4 & 38.1±21.2 & 15.3±9.7\\
\midrule
\textit{FedAvg} & \textit{Fed-MOTIF} & 68.0±9.8 & 67.5±12.0 & \bf 63.9±11.8 & 55.4±9.0 & 56.1±12.5 & \bf 37.5±11.9\\
\textit{FedAvg} & \textit{Fed-IMP} & 68.8±6.6 & 68.5±10.5 & 59.6±11.9 & \bf 59.1±8.3 & \bf 61.6±12.2 & 26.1±7.9\\
\textit{FedSGD} & \textit{Fed-MOTIF} & 34.2±12.5 & 31.0±14.1 & 61.5±14.8 & 34.8±10.4 & 33.5±12.0 & 38.7±16.4\\
\textit{FedSGD} & \textit{Fed-IMP} & 46.4±12.5 & 46.0±15.0 & 49.5±13.4 & 37.2±11.5 & 38.8±12.9 & 24.0±10.0\\
\midrule
All seen & \textit{MOTIF} & \bf 71.1±7.6 & \bf 70.6±10.3 & 57.4±12.2 & 55.4±10.8 & 56.2±15.1 & 29.9±9.7\\
All seen & \textit{IMP} & 67.8±10.8 & 67.7±14.0 & 55.2±10.5 & 58.0±8.9 & 59.9±11.9 & 21.7±7.3\\
\bottomrule
    \end{tabular}
    \caption{
    Results for the \textbf{Predicate Classification} (left) and \textbf{Scene Graph Generation} (right) tasks for Centralized and Federated Learning.
    All configurations are run 5 times using random seeds.
    }
    \label{res:rel_pred}
    \end{table*}

\subsection{Evaluation Metrics}

We follow Sanner et al.'s \cite{sanner2024voxelscenegraphintracranial} evaluation setup. The quality of detected objects is measured using Average Recall (AR) and Average Precision (AP) at a $30\%$ Intersection Over Union (IoU) threshold. 

Relation prediction is evaluated using Recall@K (R@K), mean Recall@K (mR@K), and Average Precision@K (mAP@K) again at a $30\%$ IoU threshold for object localization. Since images have on average 1 to 2 relations and up to 7, we choose to use $K=8$. All methods are evaluated for both \textbf{Predicate Classification} and \textbf{Scene Graph Generation} tasks \cite{Tang2020}, i.e. respectively predicting relations from ground truth and predicted object localization. For the latter task, we additionally give metrics upper bound given the objects that have been detected. This is especially relevant since some datasets contain significantly more small bleedings that are not contained in any relation triplet. While it is still important to be able to detect such bleedings, they are often of less clinical relevance. 
All configurations are run 5 times using random seeds.

\subsection{Centralized vs Federated Learning}

\textit{Avg. seen} refer to average results of models trained using only one dataset and evaluated on the in-distribution
test data. 
Further, we evaluate the robustness of these models by evaluating them on their corresponding other three datasets (\textit{Avg. unseen}). The expected drop in performance between the two setups corresponds to the \textbf{domain gap}. \textit{All seen} models are trained centrally using on all four datasets and provide a closer comparison to FedL. \textit{FedAvg}, and \textit{FedSGD} are trained using FedL on all four datasets.

\begin{table*}
    \centering
    \begin{tabular}{l ccccc}
    \toprule
    &\multicolumn{5}{c}{\textbf{Bleeding Detection per Type}}\\
    \cmidrule(lr){2-6}
    Method & (1)$^+$ & (2)$^{++}$ & (3)$^{++}$ & (4)$^+$ & (5)\\
    \midrule
Avg. seen & 80.6±11.6 & 39.6±16.2 & 69.0±6.8 & 83.9±12.1 & 29.6±10.7\\
Avg. unseen & 72.9±18.1 & 42.3±21.3 & 57.3±15.4 & 73.4±19.5 & 25.2±17.0\\
\midrule
\textit{FedAvg} & 87.7±6.9 & \bf 54.5±13.7 & \bf 77.5±8.6 & \bf 84.4±10.0 & \bf 35.8±13.6\\
\textit{FedSGD} & \bf 88.2±6.8 & \bf 54.5±11.9 & 76.3±8.2 & 83.5±12.9 & 35.3±16.0\\
\midrule
All seen & 87.1±7.2 & 50.6±13.3 & 70.8±8.8 & 82.9±11.7 & 34.8±14.4\\
\bottomrule
    \end{tabular}
    \caption{
    Recall for bleeding detection per type for Centralized and Federated training.
    The bleeding types refer to: 1) intraparenchymal, 2) epidural or subdural, 3) intraventricular, 4) basal subarachnoidal, and 5) non-basal subarachnoidal.
    "Basal" refers to the basal cistern, where the subarachnoidal bleeding can be more prominent.
    Superscript $^+$ denote the clinical importance of each ICH type from a surgical treatment perspective.
    All configurations are run 5 times using random seeds.
    }
    \label{res:obj_detec_type}
    \end{table*}

\begin{table*}
    \centering
    \begin{tabular}{c ll ccc ccc}
    \toprule
    \parbox[t]{3mm}{\multirow{2}{*}{\rotatebox[origin=c]{90}{Bias}}}
    &&&\multicolumn{3}{c}{\textbf{Predicate Classification}} & \multicolumn{3}{c}{\textbf{Scene Graph Generation}}\\
    \cmidrule(lr){4-6}\cmidrule(lr){7-9}
    & Method & Model & R@8$\uparrow$ & mR@8$\uparrow$ & mAP@8$\uparrow$ & R@8$\uparrow$ & mR@8$\uparrow$ & mAP@8$\uparrow$\\
    \midrule
    \parbox[t]{3mm}{\multirow{4}{*}{\rotatebox[origin=c]{90}{All datasets}}}
& Avg. seen & \textit{MOTIF} & 64.8±10.7 & 64.7±9.8 & 58.5±14.3 & 45.5±8.8 & 47.7±11.0 & 33.6±14.2\\
& Avg. seen & \textit{IMP} & 63.5±5.6 & 62.8±7.6 & 53.3±15.3 & 50.4±10.8 & 51.4±12.5 & 24.3±11.6\\
& Avg. unseen & \textit{MOTIF} & 54.4±17.5 & 54.7±19.1 & 53.9±15.7 & 33.6±18.4 & 35.6±20.8 & 23.7±15.9\\
& Avg. unseen & \textit{IMP} & 50.2±16.4 & 50.1±18.0 & 47.5±15.6 & 36.0±18.5 & 39.1±21.8 & 17.1±11.7\\
\midrule
    \midrule
    \parbox[t]{3mm}{\multirow{6}{*}{\rotatebox[origin=c]{90}{Disabled}}}
& \textit{FedAvg} & \textit{Fed-MOTIF} & 69.6±8.7 & 68.2±11.1 & \bf61.2±11.3 & 56.3±9.5 & 56.8±14.0 & \bf38.9±13.3\\
& \textit{FedAvg} & \textit{Fed-IMP} & 65.3±8.5 & 64.8±11.7 & 54.4±9.0 & \bf64.1±9.1 &\bf 65.4±11.0 & 21.3±7.1\\
& \textit{FedSGD} & \textit{Fed-MOTIF} & 46.8±16.9 & 46.4±18.9 & 53.4±14.7 & 45.7±11.2 & 43.0±15.1 & 34.1±15.4\\
& \textit{FedSGD} & \textit{Fed-IMP} & 44.5±13.4 & 47.2±14.3 & 47.7±10.4 & 40.3±10.4 & 46.6±8.4 & 9.4±5.7\\
\cmidrule{2-9}
& All seen & \textit{MOTIF} & \bf72.5±9.4 & \bf70.8±11.7 & 54.2±13.8 & 54.0±9.2 & 55.9±13.6 & 31.1±7.8\\
& All seen & \textit{IMP} & 67.9±8.0 & 68.7±9.1 & 33.9±5.3 & 63.2±8.5 & \bf65.3±11.7 & 15.8±5.2\\
\bottomrule
    \end{tabular}
    \caption{
    \textbf{Ablation study}: effect of frequency bias layers. \textit{Avg. seen} and \textit{Avg. unseen} have a bias initialized using
    statistics over all datasets. All FedL methods, and \textit{All seen} have their bias disabled.
    All configurations are run 5 times using random seeds.
    }
    \label{res:ab_bias}
    \end{table*}

\section{Results}
\label{sec:res}

In this section, we show how Federated Learning outperforms Centralized Learning first for object detection and then for relation prediction. Afterward, we perform a finer analysis with bleeding detection per ICH-type and present some qualitative results. Per-dataset results for object detection are available in the Supplementary Material.

\subsection{Centralized vs Federated Learning}
\label{sec:cl_vs_fl}


\begin{figure*}
\centering

  \begin{subfigure}{0.49\linewidth}
    \centering
    \includegraphics[width=\linewidth]{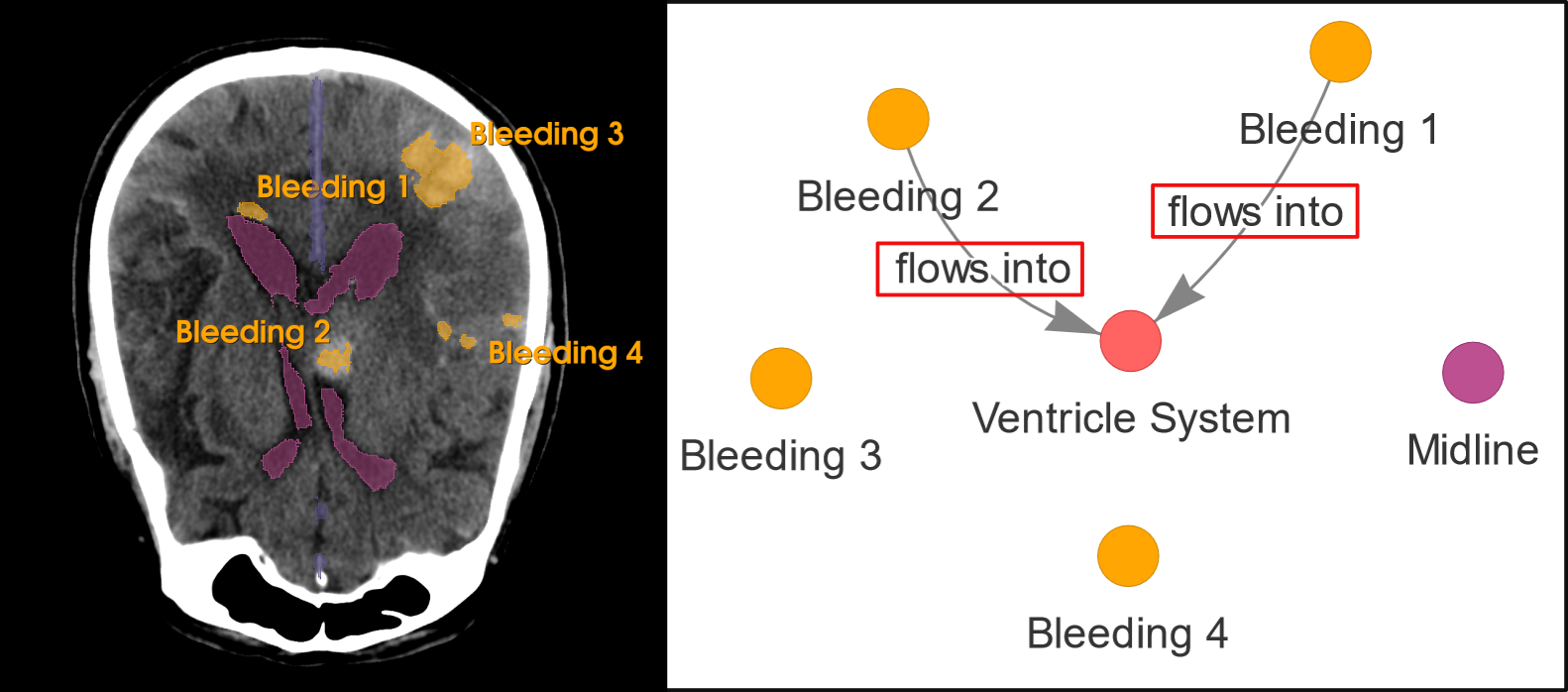}
    \caption{Model trained on the \textbf{INST} dataset (out-of-distribution)}
  \end{subfigure}
    \hfill
  \begin{subfigure}{0.49\linewidth}
    \centering
    \includegraphics[width=\linewidth]{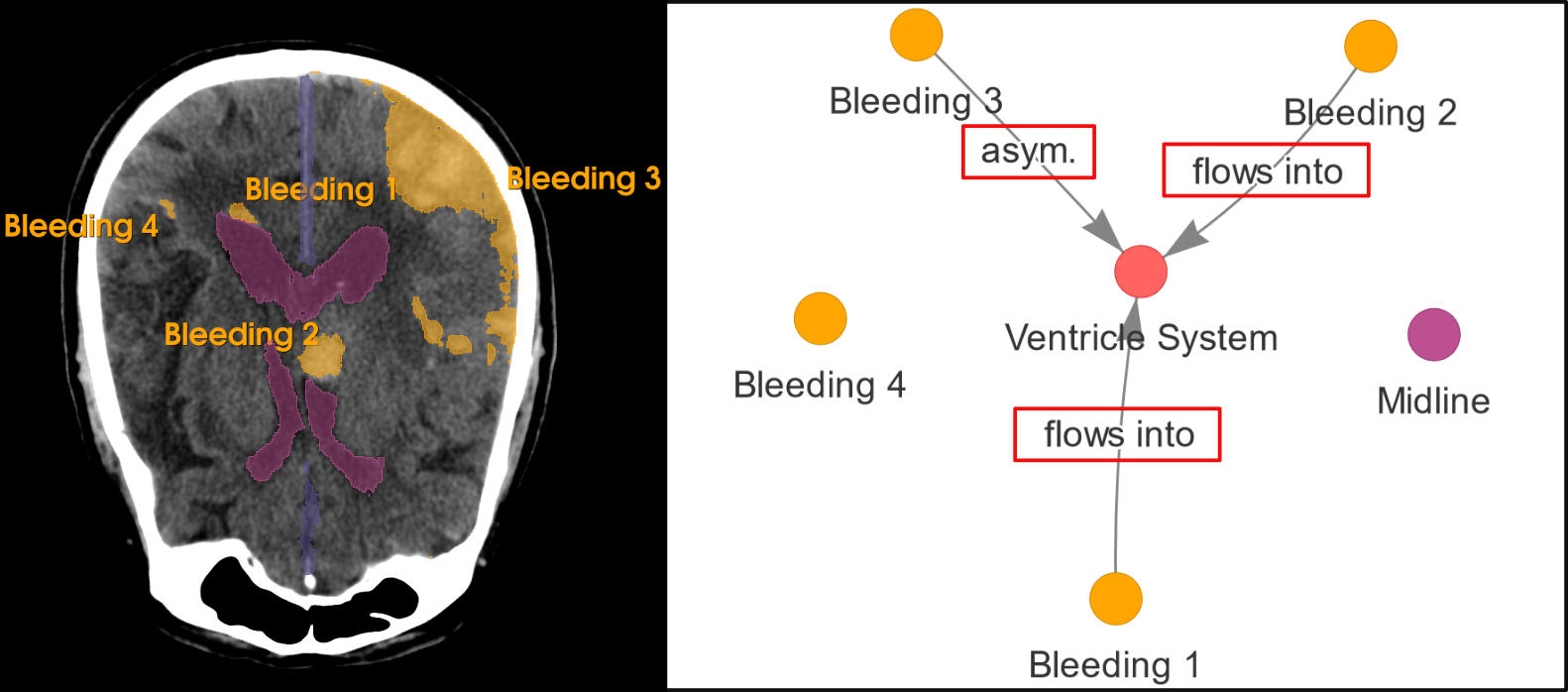}
    \caption{Model trained on \textbf{all datasets} using \textbf{\textit{FedL}}}
  \end{subfigure}
  
\caption{
Qualitative results: predicted segmentation and relations for a \textbf{patient from the BHSD dataset} by \textit{MOTIF} trained on the INST dataset (left) and using \textit{Fed-MOTIF} (right). Both models detect the intraventricular \textcolor{Dandelion}{bleedings 1 and 2} and that they are in the \textcolor{Ventricle}{ventricle system} through the corresponding relation. The model trained on INST already provides a worse segmentation of \textcolor{Dandelion}{bleeding 2}. It also vastly undersegments \textcolor{Dandelion}{bleeding 3} to the point where the same ICH instance also gets detected as an additional \textcolor{Dandelion}{bleedings 4}. The model trained with \textit{FedAvg} localizes \textcolor{Dandelion}{bleeding 3} much more precisely and detects correctly a \textcolor{Ventricle}{ventricle system} asymmetry. Additionally, a small subarachnoidal \textcolor{Dandelion}{bleeding 4} is only detected by this model (right). Though, both models fail to detect a \textcolor{Midline}{midline} shift.
}

\label{fig:qual}
\end{figure*}

\subsubsection{Object Detection}

The \textit{Retina-UNet} architecture can detect objects using two distinct mechanisms. The results are shown in \cref{res:obj_detec}.

\paragraph{Ventricle System \& Midline Detection:} These two anatomies are detected using semantic segmentation rather than traditional object detection. This approach provides a precise localization for in-distribution data, but can occasionally fail on out-of-distribution data. In particular, the models trained on the PC dataset fail to generalize, which can be partially explained by the frequent occurrence of intraventricular bleeding in a small dataset. This limits the ventricular volume available to learn to segment the anatomy properly. Similarly, large portions of the midline can also contain bleeding and lead to the learning of spurious correlation within the training data. In contrast, all models trained using all datasets, whether trained centrally or using FedL, can detect both anatomies at a $90$+\% rate, as shown in \cref{res:obj_detec}.

\paragraph{Bleeding Detection:} With the high variability in ICH manifestation, the results of \textit{Avg. unseen} are both worse and more variable compared to in-distribution \textit{Avg. seen}. This is particularly striking when considering the upper bounds for relation predictions in \cref{res:obj_detec} (right). It indicates that \textbf{the models trained using only one dataset fail to reliably detect clinically relevant ICH}. Here, FedL shows its full potential and allows the training of models on par, or even out-performing models trained centrally on all datasets. This phenomenon can be attributed either to 1) a rebalancing of the dataset weights (equal using FedL, dataset length using centralized learning) or 2) FedL simulating a higher batch size, leading to better model convergence.
Overall, \textbf{the models trained using FedL can reliably detect relevant bleedings across datasets}.


\subsubsection{Relation Prediction}

Relations can be predicted under two setups, depending on whether ground truth object localization or predicted ones are used. Results for both tasks are available in \cref{res:rel_pred}.

\paragraph{Predicate Classification:} Once again, the models trained centrally perform decently on in-distribution data, but already perform 3\% to 8\% worse, \cref{res:rel_pred} (left). There is of course a shift in relation distribution, but the burden of generalization for relation prediction is double. The Predicate Classification task is constructed to evaluate relation prediction independently of the prediction quality of the object detector, i.e. by using ground truth object localization. However, the features generated by the trained detector are still used for relation prediction. As such, \textbf{if a detector fails on unseen data, one can assume that the features generated on unseen data are also disadvantageous for relation prediction}. This issue again calls for improved model generalizability as a whole. On the other hand, \textbf{models trained using FedL perform reliably on all datasets}.

\paragraph{Scene Graph Generation:} Compared to the previous task, this one showcases real-world performance of the entire prediction pipeline. The models trained on single datasets see their performance drop significantly even on in-distribution data when using predicted object localization. Additionally, the performance loss on unseen data reaches up to an additional $8$\%, \cref{res:rel_pred} (right). 
\textbf{Models trained with FedL bridge the domain gap by achieving a performance on par with models trained centrally with all datasets}.

\subsection{Result Analysis}

\subsubsection{Bleeding-type-based Detection}

ICH manifestation is vastly diverse, but it can be categorized based on anatomical location. \cref{res:obj_detec_type} offers further insights into ICH detection performance. For instance, basal subarachnoidal bleedings are tiny and thus harder to detect. However, these are never part of relation triplets. In contrast, \textbf{intraventricular bleedings crucial to detect}, as they always have a relation with the ventricle system by design. \textbf{Both \textit{FedAvg}, and \textit{FedSGD} enable models to detect significantly more such bleedings compared to models trained centrally}. Similarly, subdural bleedings are often associated with a midline shift or an asymmetrical ventricle system. FedL methods again offer superior performance. 

\subsubsection{Ablation Study: Relation Bias Importance}


To evaluate the impact of the frequency bias layer on model performance, we consider two setups: 1) we train the models centrally using one dataset, but the bias layer is initialized using statistics over all datasets, and 2) we train using all datasets using either centrally or using FedL, but with the bias layer disabled.

The models trained on a single dataset (first four rows in \cref{res:ab_bias}) still have a similar performance on in-distribution data. However, they show a slight improvement in robustness on unseen data. Models trained on all datasets without a bias layer, whether centrally or with FedL, perform similarly. 
These results show that learning adequate domain statistics over the data can improve model generalizability when training data is limited. 


\subsubsection{Qualitative Results}

We evaluate method performance qualitatively for a hard clinical case. \cref{fig:qual} shows how the different training setups influence the detection precision for multiple ICH types, as well as the segmentation quality.
These Scene Graphs can easily help clinicians prioritize the treatment of patients in critical condition by summarizing a patient's state.

\section{Conclusion}

Intracranial Hemorrhage (ICH) is a critical condition, which can manifest in numerous ways and shift across clinical centers worldwide.
We pioneer Federated Voxel Scene Graph Generation and train more robust models, which generalize across multiple datasets without needing to share patient data.
While pure ICH detection only provides a superficial understanding of the clinical cerebral scene, our method learns to model complex relations between ICH and adjacent brain structures.
We evaluate our method on four datasets against centralized learning methods. 
We demonstrate how models trained on a single centralized dataset fail to bridge the domain gap on shifted data.
In contrast, models trained using our method can recall up to $20$\% more clinically relevant relations for Scene Graph Generation and can better support the clinical decision-making. 
Only if models can detect the relations in such diverse cases, will we have achieved progress towards usable Deep-Learning solutions for clinical applications.

\section{Compliance with Ethical Standards}

This study was performed in line with the principles of the Declaration of Helsinki. The retrospective evaluation of imaging data from the University Medical Center Mainz was approved by the local ethics boards (Project 2021-15948-retrospektiv).

{\small
\bibliographystyle{ieee_fullname}
\bibliography{sources}
}

\clearpage
\section{Supplementary Material}
\label{sec:sup_mat}

\begin{figure*}
  \centering
  
  \begin{subfigure}[t]{0.33\linewidth}
    \centering
    \includegraphics[width=\linewidth]{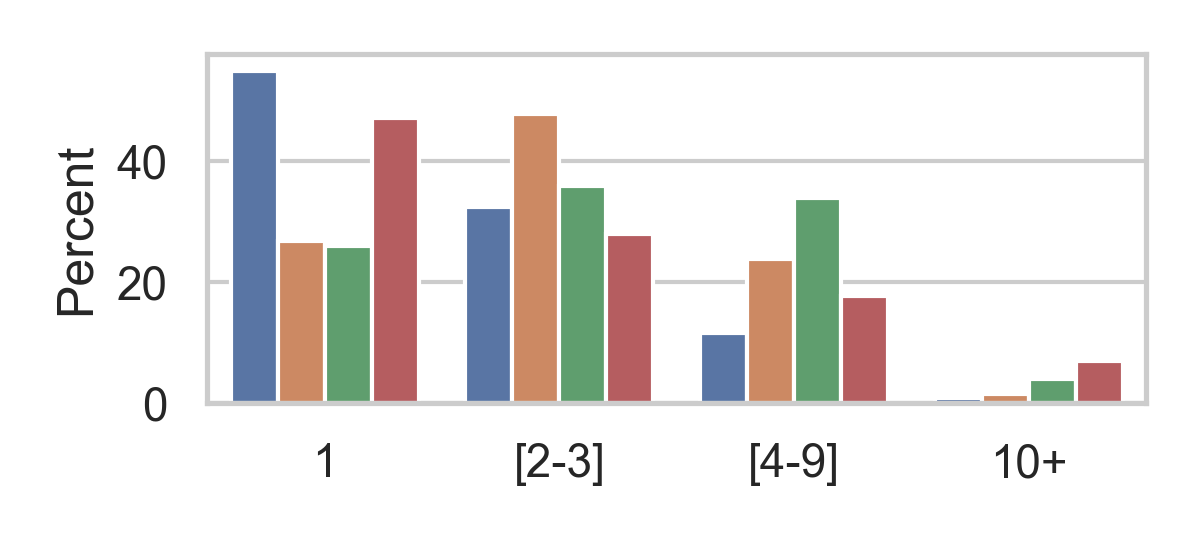}
    \caption{Number of bleedings per image}
  \end{subfigure}
  \hfill
  \begin{subfigure}[t]{0.33\linewidth}
    \centering
    \includegraphics[width=\linewidth]{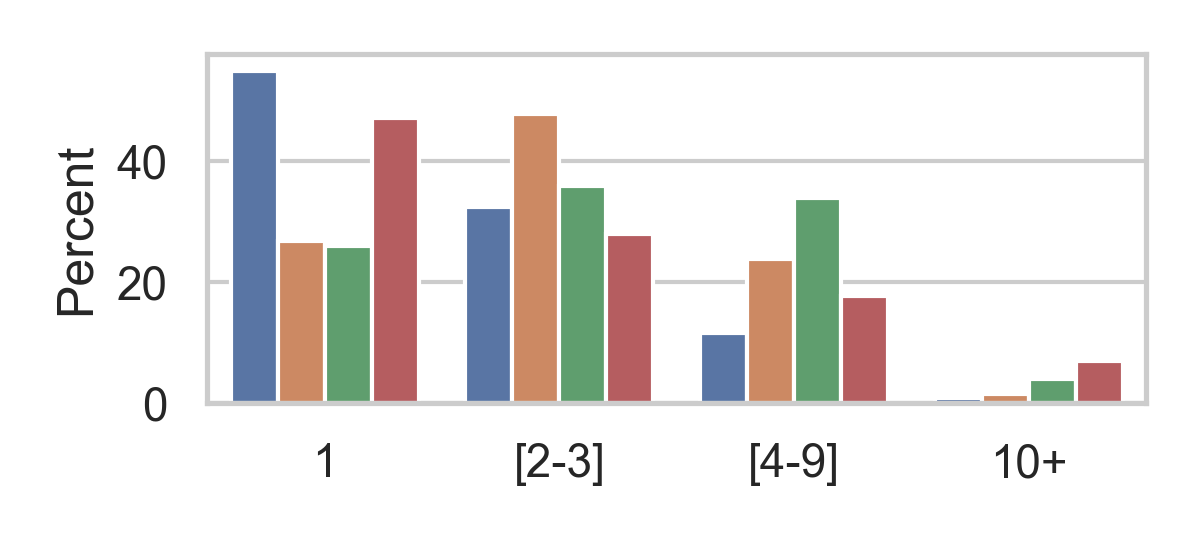}
    \caption{Bleeding volume distribution (cm3)}
  \end{subfigure}
  \hfill
  \begin{subfigure}[t]{0.33\linewidth}
    \centering
    \includegraphics[width=\linewidth]{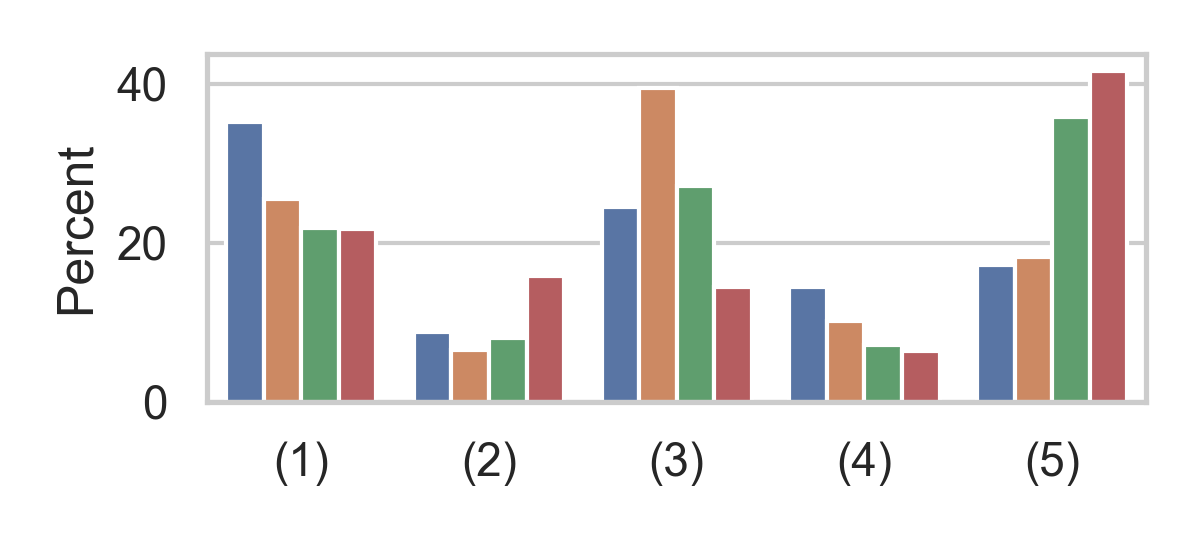}
    \caption{Bleeding type distribution}
  \end{subfigure}  
  
  \begin{subfigure}[t]{0.33\linewidth}
    \centering
    \includegraphics[width=\linewidth]{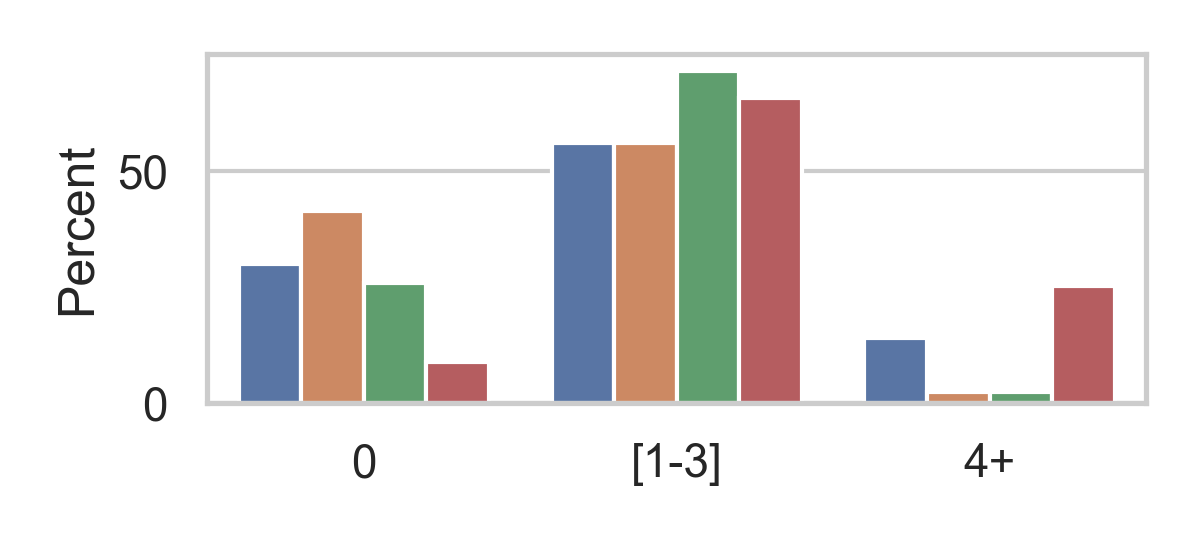}
    \caption{Number of relations per image}
  \end{subfigure}
  \hfill
  \begin{subfigure}[t]{0.33\linewidth}
    \centering
    \includegraphics[width=\linewidth]{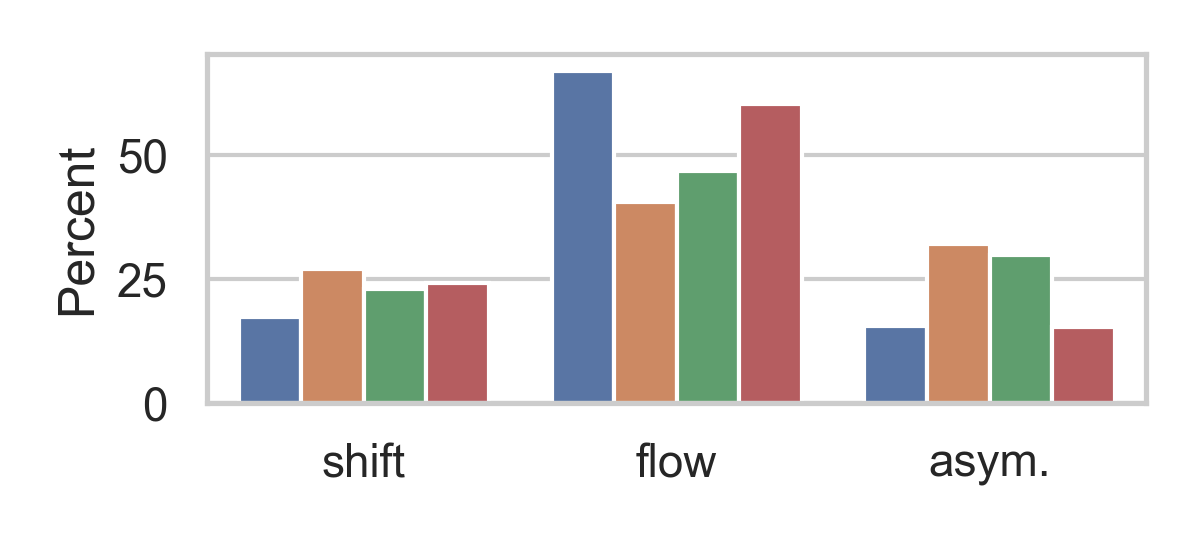}
    \caption{Relation distribution}
  \end{subfigure}
  \hfill
  \begin{subfigure}[t]{0.33\linewidth}
    \centering
    \raisebox{3mm}{\includegraphics[width=.5\linewidth]{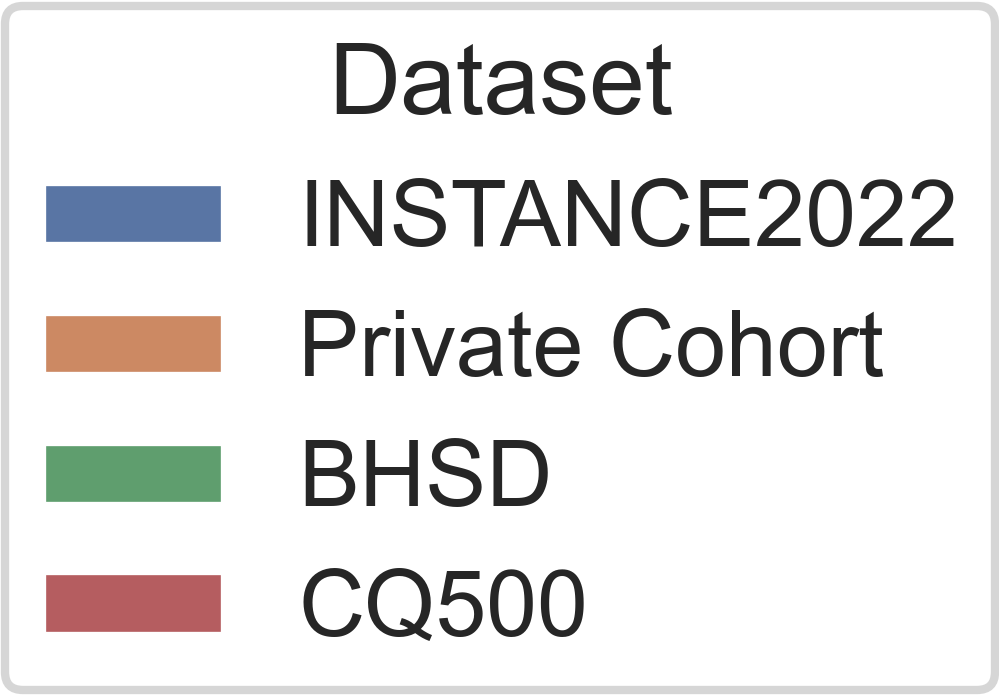}}
  \end{subfigure}
  
  \caption{
Distribution of bleedings and relations for each dataset. All datasets show in bleeding representation whether regarding their number, volume or type. The bleeding types refer to: 1) intraparenchymal, 2) epidural or subdural, 3) intraventricular, 4) basal subarachnoidal, and 5) non-basal subarachnoidal. "Basal" refers to the basal cistern, where the subarachnoidal bleeding can be more prominent.
  }
\label{fig:data_stats}
\end{figure*}

\subsection{Data Preprocessing}

We harmonize the slice thickness of all images such that all the relevant anatomy fits in a $32\times512\times512$ volume. 
The resampling of thick-sliced head CTs to thicker slices will create interpolation artifacts along the skull, if the new thickness is not a multiple of the original one. As such, we resample images with a slice-thickness of $2.5$ mm to $5.0$ mm. 
Images with a slice thickness above $4.0$ mm are not resampled. As some images extend up to the patient's shoulders, we crop volumes with more than 32 slices. 
Since no masks are available for CQ500 and PC, we use an \textit{nnUNet} model \cite{Isensee2021} pre-trained on INST to produce a pre-segmentation of all ICH. 

\subsection{Model Training}
We use the official data split released with the INST and BHSD datasets. Due to our data curation process, our results are not directly comparable to other publications. Evaluating our models on the original data splits is also impossible, as the annotation does not exist for images which have not been selected.  
As some cases do not contain any relations, we only keep images with relations for relation detection. Split sizes can be found in (\cref{sup:split_detec,sup:split_rel}). Additionally, given the number and length of experiments (7h for object detection and 1h for relation prediction), it is not feasible to find optimal hyperparameters for all setups. As such, we use the same hyperparameters for centralized and federated experiments.
Exact splits and detailed configuration files will be made available along the configuration files.

\begin{table}[ht]
\centering
\begin{tabular}{l c c c}
\toprule
Dataset & Training & Validation & Test\\
\midrule
INSTANCE2022 & 81 & 12 & 27\\
Private Cohort & 41 & 7 & 19\\
BHSD & 51 & 10 & 39\\
CQ500 & 86 & 14 & 57\\
\bottomrule
\end{tabular}
\caption{
Split sizes for object detection experiments.
}
\label{sup:split_detec}
\end{table}

\begin{table}[ht]
\centering
\begin{tabular}{l c c c}
\toprule
Dataset & Training & Validation & Test\\
\midrule
INSTANCE2022 & 56 & 9 & 24\\
Private Cohort & 38 & 7 & 16\\
BHSD & 35 & 6 & 29\\
CQ500 & 51 & 11 & 30\\
\bottomrule
\end{tabular}
\caption{
Split sizes for relation prediction experiments.
}
\label{sup:split_rel}
\end{table}

\subsection{Implementation Details}
\label{sec:impl}

The methods are implemented in Python 3.10 and PyTorch 2.4 \cite{paszke2019pytorch} and make use of the Voxel Scene Graph Generation framework \cite{sanner2024voxelscenegraphintracranial} and of the open-source framework \textit{TheODen}\footnote{\theoden} as an overlay to enable federated training. The federated training is not simulated, as each of the 4 clients had an own computer with an NVIDIA RTX 4090 GPU. An additional fifth computer is used for model aggregation and does not require a GPU. For each federated training, we train for 16000 and 200 steps respectively for object detection and relation prediction. Each training round lasts 25 and 5 steps respectively and is followed by an aggregation round. The V-RAM requirement was optimized to use the GPUs' full 24 GB of memory and allows for a batch size of 1 and 13 respectively for object detection and relation prediction.

\subsection{Detailed Results}

The tables that follow provide results for each dataset separately.


\begin{table}[ht]
        \centering
        \begin{tabular}{l cccc}
        \toprule
        &\multicolumn{4}{c}{Ventricle System Segmentation}\\
        \cmidrule(lr){2-5}
        Train & INST & PC & BHSD & CQ500\\
        \cmidrule(lr){1-1}\cmidrule(lr){2-2}\cmidrule(lr){3-3}\cmidrule(lr){4-4}\cmidrule(lr){5-5}
INST & 73.8±2.0 & 65.3±2.8 & 61.0±6.5 & 60.0±9.3 \\
PC & 56.2±7.1 & 80.4±2.3 & 28.9±9.5 & 16.9±6.1 \\
BHSD & \bf78.5±1.0 & 69.3±4.3 & \bf78.3±1.1 & \bf78.1±1.0 \\
CQ500 & 77.2±1.0 & 71.0±3.3 & 76.0±2.1 & \bf78.2±2.1 \\
\midrule
\textit{FedAvg} & 78.2±0.4 & \bf81.2±1.0 & 72.8±2.9 & 74.6±3.5 \\
\textit{FedSGD} & 77.3±0.6 & 79.6±0.9 & 72.0±2.9 & 73.6±2.6 \\
\midrule
all & 77.0±0.6 & 79.1±0.9 & 72.9±3.3 & 74.9±2.1 \\
\bottomrule
        \end{tabular}
        \caption{
\textbf{Patient Dice score} for the \textbf{ventricle system}, when training in a centralized setup using one or all datasets or using FedL.
}
    \end{table}

\begin{table}[ht]
        \centering
        \begin{tabular}{l cccc}
        \toprule
        &\multicolumn{4}{c}{Midline Segmentation}\\
        \cmidrule(lr){2-5}
        Train & INST & PC & BHSD & CQ500\\
        \cmidrule(lr){1-1}\cmidrule(lr){2-2}\cmidrule(lr){3-3}\cmidrule(lr){4-4}\cmidrule(lr){5-5}
INST & 65.5±1.6 & 50.1±2.8 & 50.1±3.3 & 48.3±4.4 \\
PC & 55.1±5.9 & 72.1±1.0 & 28.1±8.2 & 22.1±7.6 \\
BHSD & 72.0±1.2 & 55.7±4.6 & 67.6±2.5 & 66.8±2.3 \\
CQ500 & 72.5±0.6 & 58.4±2.3 & 67.7±0.9 & 67.2±2.5 \\
\midrule
\textit{FedAvg} & \bf73.6±0.5 & \bf70.1±0.8 & \bf68.4±1.1 & \bf67.8±1.6 \\
\textit{FedSGD} & 70.4±1.5 & 64.9±1.1 & 62.1±2.7 & 60.6±2.6 \\
\midrule
all & 71.7±0.4 & 64.5±0.9 & 64.8±1.9 & 62.4±2.0 \\
\bottomrule
        \end{tabular}
        \caption{
\textbf{Patient Dice score} for the \textbf{midline}, when training in a centralized setup using one or all datasets or using FedL.
}
    \end{table}

\begin{table}[ht]
        \centering
        \begin{tabular}{l cccc}
        \toprule
        &\multicolumn{4}{c}{Bleeding Segmentation}\\
        \cmidrule(lr){2-5}
        Train & INST & PC & BHSD & CQ500\\
        \cmidrule(lr){1-1}\cmidrule(lr){2-2}\cmidrule(lr){3-3}\cmidrule(lr){4-4}\cmidrule(lr){5-5}
INST & 79.1±0.7 & 81.8±0.3 & 57.2±1.2 & 40.5±2.5 \\
PC & 62.0±4.7 & \bf83.1±0.6 & 29.4±5.0 & 19.0±3.6 \\
BHSD & 74.2±0.6 & 79.0±0.9 & 65.5±0.8 & 49.8±1.9 \\
CQ500 & 70.0±2.0 & 73.4±1.0 & 61.3±1.1 & 52.7±1.3 \\
\midrule
\textit{FedAvg} & \bf81.0±0.4 & 82.6±0.5 & \bf70.1±0.5 & \bf55.3±1.6 \\
\textit{FedSGD} & 79.6±0.7 & 81.4±0.6 & 68.2±0.9 & 53.4±1.4 \\
\midrule
all & 78.3±0.8 & 80.0±0.7 & 67.3±0.8 & 53.5±1.2 \\
\bottomrule
        \end{tabular}
        \caption{
\textbf{Patient Dice score} for \textbf{bleeding}, when training in a centralized setup using one or all datasets or using FedL.
}
    \end{table}

\begin{table*}
        \centering
        \begin{tabular}{ll ccc ccc}
        \toprule
        &&\multicolumn{6}{c}{\textbf{INST Dataset}}\\
        \cmidrule(lr){3-8}
        &&\multicolumn{3}{c}{\textbf{Predicate Classification}} & \multicolumn{3}{c}{\textbf{Scene Graph Generation}}\\
        \cmidrule(lr){3-5}\cmidrule(lr){6-8}
        Method & Model & R@8$\uparrow$ & mR@8$\uparrow$ & mAP@8$\uparrow$ & R@8$\uparrow$ & mR@8$\uparrow$ & mAP@8$\uparrow$\\
        \midrule
INST dataset & \textit{MOTIF} & 62.9±5.0 & 65.1±5.7 & 57.2±6.8 & 47.1±6.1 & 48.7±6.3 & 38.0±5.9\\
INST dataset & \textit{IMP} & 61.5±3.2 & 65.0±2.8 & 53.9±3.4 & 58.1±4.6 & 60.1±3.9 & 24.7±5.4\\
Avg. unseen & \textit{MOTIF} & 63.6±16.3 & 65.0±15.5 & 62.8±8.8 & 46.8±16.8 & 46.4±15.6 & 38.7±15.8\\
Avg. unseen & \textit{IMP} & 61.7±12.0 & 63.8±11.2 & 55.8±11.4 & 52.5±14.6 & 54.0±14.5 & 25.9±7.0\\
\midrule
\textit{FedAvg} & \textit{Fed-MOTIF} & \bf 76.9±4.6 & \bf 77.8±4.9 &  \bf 70.5±2.0 & 60.7±6.0 & 58.3±6.9 & \bf 45.3±8.2\\
\textit{FedAvg} & \textit{Fed-IMP} & 73.1±3.1 & 73.8±3.3 & 65.5±5.3 &  \bf 65.1±3.4 &  \bf 65.8±3.6 & 29.6±1.1\\
\textit{FedSGD} & \textit{Fed-MOTIF} & 43.3±9.6 & 42.7±10.5 & 69.2±8.1 & 46.7±6.5 & 43.7±3.7 & 57.1±6.0\\
\textit{FedSGD} & \textit{Fed-IMP} & 54.5±1.8 & 55.3±2.2 & 50.6±8.0 & 46.0±4.6 & 44.2±4.2 & 31.0±2.7\\
\midrule
All seen & \textit{MOTIF} & 74.3±1.5 & 76.1±2.4 & 60.2±4.4 & 57.6±6.1 & 57.6±5.9 & 37.1±2.2\\
All seen & \textit{IMP} & 70.7±1.9 & 73.8±1.0 & 61.2±2.3 & 60.3±5.7 & 62.2±5.1 & 27.9±3.0\\
\bottomrule
        \end{tabular}
        \caption{
        Results for the \textbf{Predicate Classification} (left) and \textbf{Scene Graph Generation} (right) tasks for Centralized and Federated Learning on the \textbf{INST} dataset). All configurations are run 5 times using random seeds.
        }
        \end{table*}
        
\begin{table*}
        \centering
        \begin{tabular}{ll ccc ccc}
        \toprule
        &&\multicolumn{6}{c}{\textbf{PC Dataset}}\\
        \cmidrule(lr){3-8}
        &&\multicolumn{3}{c}{\textbf{Predicate Classification}} & \multicolumn{3}{c}{\textbf{Scene Graph Generation}}\\
        \cmidrule(lr){3-5}\cmidrule(lr){6-8}
        Method & Model & R@8$\uparrow$ & mR@8$\uparrow$ & mAP@8$\uparrow$ & R@8$\uparrow$ & mR@8$\uparrow$ & mAP@8$\uparrow$\\
        \midrule
PC dataset & \textit{MOTIF} & 76.6±6.5 & 80.6±4.7 & 78.3±2.5 & 58.3±4.3 & 64.0±3.7 & 51.1±8.9\\
PC dataset & \textit{IMP} & 63.8±7.3 & 66.1±9.0 & 73.8±5.0 & 57.7±8.2 & 59.1±4.7 & 21.0±7.4\\
Avg. unseen & \textit{MOTIF} & 67.9±14.3 & 70.7±16.2 & 74.8±4.7 & 53.1±11.7 & 60.6±11.9 & 40.8±9.9\\
Avg. unseen & \textit{IMP} & 68.2±9.6 & 74.0±9.5 & 67.7±9.5 & 55.2±10.5 & 63.7±11.1 & 26.1±7.7\\
\midrule
\textit{FedAvg} & \textit{Fed-MOTIF} & 72.8±5.6 & 78.6±5.9 &  \bf76.7±5.2 & 64.2±1.6 & 71.7±1.0 &  \bf46.9±5.2\\
\textit{FedAvg} & \textit{Fed-IMP} & 71.9±4.2 & 80.3±4.1 & 73.3±8.1 & 65.4±3.0 & 75.1±3.9 & 34.2±4.1\\
\textit{FedSGD} & \textit{Fed-MOTIF} & 39.1±9.8 & 38.6±13.4 & 72.1±3.8 & 39.7±5.4 & 45.5±5.1 & 45.5±11.3\\
\textit{FedSGD} & \textit{Fed-IMP} & 60.6±3.6 & 64.7±5.2 & 66.5±2.7 & 46.7±8.8 & 53.2±9.6 & 32.7±6.9\\
\midrule
All seen & \textit{MOTIF} & 78.5±1.1 & 83.7±1.6 & 72.6±3.7 &  \bf70.0±5.7 &  \bf78.2±3.5 & 37.7±6.4\\
All seen & \textit{IMP} & \bf 81.0±5.7 & \bf 86.9±4.3 & 67.6±4.2 & 69.8±3.6 & 77.1±2.0 & 28.0±4.8\\
\bottomrule
        \end{tabular}
        \caption{
        Results for the \textbf{Predicate Classification} (left) and \textbf{Scene Graph Generation} (right) tasks for Centralized and Federated Learning on the \textbf{PC} dataset). All configurations are run 5 times using random seeds.
        }
        \end{table*}
        
\begin{table*}
        \centering
        \begin{tabular}{ll ccc ccc}
        \toprule
        &&\multicolumn{6}{c}{\textbf{BHSD Dataset}}\\
        \cmidrule(lr){3-8}
        &&\multicolumn{3}{c}{\textbf{Predicate Classification}} & \multicolumn{3}{c}{\textbf{Scene Graph Generation}}\\
        \cmidrule(lr){3-5}\cmidrule(lr){6-8}
        Method & Model & R@8$\uparrow$ & mR@8$\uparrow$ & mAP@8$\uparrow$ & R@8$\uparrow$ & mR@8$\uparrow$ & mAP@8$\uparrow$\\
        \midrule
BHSD dataset & \textit{MOTIF} & 67.9±2.5 & 58.1±4.2 & 36.6±2.3 & 38.3±5.9 & 37.3±8.4 & 17.7±3.5\\
BHSD dataset & \textit{IMP} & 61.7±6.3 & 52.2±6.0 & 34.9±5.9 & 38.8±3.7 & 35.2±4.9 & 6.9±2.1\\
Avg. unseen & \textit{MOTIF} & 51.1±17.7 & 41.0±16.3 & 41.8±10.7 & 31.6±14.9 & 25.9±14.0 & 15.7±9.7\\
Avg. unseen & \textit{IMP} & 51.7±15.8 & 40.5±12.9 & 39.1±9.5 & 31.6±15.3 & 30.1±15.5 & 9.7±6.8\\
\midrule
\textit{FedAvg} & \textit{Fed-MOTIF} & 68.0±5.9 & 54.4±5.9 & 47.8±5.7 & \bf 47.1±4.0 & 39.8±2.6 &  \bf20.9±5.3\\
\textit{FedAvg} & \textit{Fed-IMP} & 69.8±6.9 & 53.9±4.6 & 48.1±5.0 & 46.5±3.0 & 43.4±5.2 & 14.8±4.1\\
\textit{FedSGD} & \textit{Fed-MOTIF} & 32.4±10.2 & 20.9±7.4 &  \bf50.8±17.9 & 29.0±5.6 & 19.7±5.1 & 17.4±2.4\\
\textit{FedSGD} & \textit{Fed-IMP} & 37.4±7.2 & 29.8±3.0 & 34.5±7.4 & 26.7±6.6 & 29.2±8.9 & 10.9±3.1\\
\midrule
All seen & \textit{MOTIF} & \bf 72.4±4.1 &  \bf59.9±5.8 & 42.4±5.3 & \bf 47.2±3.8 & 39.8±6.1 & 17.1±3.7\\
All seen & \textit{IMP} & 66.2±5.1 & 54.4±5.3 & 47.2±5.7 & 49.8±3.3 &  \bf 48.1±3.7 & 12.4±0.9\\
\bottomrule
        \end{tabular}
        \caption{
        Results for the \textbf{Predicate Classification} (left) and \textbf{Scene Graph Generation} (right) tasks for Centralized and Federated Learning on the \textbf{BHSD} dataset). All configurations are run 5 times using random seeds.
        }
        \end{table*}
        
\begin{table*}
        \centering
        \begin{tabular}{ll ccc ccc}
        \toprule
        &&\multicolumn{6}{c}{\textbf{CQ500 Dataset}}\\
        \cmidrule(lr){3-8}
        &&\multicolumn{3}{c}{\textbf{Predicate Classification}} & \multicolumn{3}{c}{\textbf{Scene Graph Generation}}\\
        \cmidrule(lr){3-5}\cmidrule(lr){6-8}
        Method & Model & R@8$\uparrow$ & mR@8$\uparrow$ & mAP@8$\uparrow$ & R@8$\uparrow$ & mR@8$\uparrow$ & mAP@8$\uparrow$\\
        \midrule
CQ500 dataset & \textit{MOTIF} & 51.5±7.7 & 55.0±7.3 & 59.0±5.3 & 34.8±3.4 & 38.6±3.9 & 31.0±5.4\\
CQ500 dataset & \textit{IMP} & 53.0±7.5 & 53.9±7.8 & 52.3±7.1 & 38.5±2.7 & 37.9±5.0 & 20.0±7.1\\
Avg. unseen & \textit{MOTIF} & 42.0±14.3 & 45.5±15.9 & 51.0±9.2 & 30.1±15.5 & 32.6±16.7 & 24.2±12.6\\
Avg. unseen & \textit{IMP} & 44.7±12.3 & 47.1±14.0 & 41.8±10.6 & 34.4±18.0 & 35.3±18.2 & 16.7±8.6\\
\midrule
\textit{FedAvg} & \textit{Fed-MOTIF} & 54.4±3.2 & 59.3±3.8 &  \bf60.3±4.0 & 49.7±7.8 & 54.6±7.5 &  \bf36.9±4.6\\
\textit{FedAvg} & \textit{Fed-IMP} & \bf 60.5±1.1 &  \bf66.0±2.1 & 51.4±4.5 & \bf 59.4±3.9 & \bf 61.9±2.4 & 25.8±2.4\\
\textit{FedSGD} & \textit{Fed-MOTIF} & 22.1±8.8 & 21.7±8.6 & 54.0±11.4 & 23.8±2.8 & 25.3±2.7 & 34.8±7.4\\
\textit{FedSGD} & \textit{Fed-IMP} & 32.9±4.9 & 34.3±4.9 & 46.5±8.1 & 29.5±7.4 & 28.8±6.6 & 21.4±5.6\\
\midrule
All seen & \textit{MOTIF} & 59.1±1.9 & 62.8±2.0 & 54.3±7.5 & 46.7±4.7 & 49.1±4.3 & 27.7±5.9\\
All seen & \textit{IMP} & 53.2±1.5 & 55.6±1.8 & 44.7±4.4 & 52.2±3.7 & 52.1±4.1 & 18.5±2.4\\
\bottomrule
        \end{tabular}
        \caption{
        Results for the \textbf{Predicate Classification} (left) and \textbf{Scene Graph Generation} (right) tasks for Centralized and Federated Learning on the \textbf{CQ500} dataset). All configurations are run 5 times using random seeds.
        }
        \end{table*}

\begin{table*}
        \centering
        \begin{tabular}{ll ccc ccc}
        \toprule
        &&\multicolumn{6}{c}{\textbf{INST Dataset}}\\
        \cmidrule(lr){3-8}
        &&\multicolumn{3}{c}{\textbf{Predicate Classification}} & \multicolumn{3}{c}{\textbf{Scene Graph Generation}}\\
        \cmidrule(lr){3-5}\cmidrule(lr){6-8}
        Method & Model & R@8$\uparrow$ & mR@8$\uparrow$ & mAP@8$\uparrow$ & R@8$\uparrow$ & mR@8$\uparrow$ & mAP@8$\uparrow$\\
        \midrule
INST dataset & \textit{MOTIF} & 62.9±5.0 & 65.1±5.7 & 57.2±6.8 & 59.7±1.4 & 61.5±2.4 & 55.5±6.2\\
INST dataset & \textit{IMP} & 61.5±3.2 & 65.0±2.8 & 53.9±3.4 & 57.2±6.5 & 60.6±5.3 & 46.7±6.6\\
Avg. unseen & \textit{MOTIF} & 63.6±16.3 & 65.0±15.5 & 62.8±8.8 & 63.0±11.5 & 62.1±9.8 & 50.8±15.3\\
Avg. unseen & \textit{IMP} & 60.9±11.6 & 63.0±10.7 & 54.3±11.2 & 57.7±9.5 & 60.3±8.3 & 46.2±11.2\\
\midrule
\textit{FedAvg} & \textit{Fed-MOTIF} & \bf 76.9±4.6 & \bf 77.8±4.9 & \bf 70.5±2.0 & \bf 75.1±1.2 & \bf 71.8±1.5 & 60.0±5.8\\
\textit{FedAvg} & \textit{Fed-IMP} & 73.1±3.1 & 73.8±3.3 & 65.5±5.3 & 70.6±3.1 & 71.3±2.6 & 56.8±2.8\\
\textit{FedSGD} & \textit{Fed-MOTIF} & 43.3±9.6 & 42.7±10.5 & 69.2±8.1 & 59.9±1.9 & 57.4±2.1 & \bf 63.3±7.0\\
\textit{FedSGD} & \textit{Fed-IMP} & 54.5±1.8 & 55.3±2.2 & 50.6±8.0 & 54.9±5.0 & 58.0±3.9 & 48.0±5.8\\
\midrule
All seen & \textit{MOTIF} & 74.3±1.5 & 76.1±2.4 & 60.2±4.4 & 71.3±2.6 & \bf  71.5±2.7 & 54.5±2.9\\
All seen & \textit{IMP} & 66.4±3.6 & 69.2±2.9 & 52.0±2.6 & 61.3±5.0 & 64.2±3.9 & 46.8±5.5\\
\bottomrule
        \end{tabular}
        \caption{
        Results for the \textbf{Predicate Classification} (left) and \textbf{Scene Graph Generation} (right) tasks for Centralized and Federated Learning on the \textbf{INST} dataset). All configurations are run 5 times using random seeds.
        }
        \end{table*}
        
\begin{table*}
        \centering
        \begin{tabular}{ll ccc ccc}
        \toprule
        &&\multicolumn{6}{c}{\textbf{PC Dataset}}\\
        \cmidrule(lr){3-8}
        &&\multicolumn{3}{c}{\textbf{Predicate Classification}} & \multicolumn{3}{c}{\textbf{Scene Graph Generation}}\\
        \cmidrule(lr){3-5}\cmidrule(lr){6-8}
        Method & Model & R@8$\uparrow$ & mR@8$\uparrow$ & mAP@8$\uparrow$ & R@8$\uparrow$ & mR@8$\uparrow$ & mAP@8$\uparrow$\\
        \midrule
PC dataset & \textit{MOTIF} & 76.6±6.5 & 80.6±4.7 & 78.3±2.5 & 61.0±3.1 & 66.9±2.4 & 66.6±5.2\\
PC dataset & \textit{IMP} & 63.8±7.3 & 66.1±9.0 & 73.8±5.0 & 58.7±5.2 & 59.7±4.3 & 48.1±10.1\\
Avg. unseen & \textit{MOTIF} & 67.9±14.3 & 70.7±16.2 & 74.8±4.7 & 60.2±8.7 & 69.9±8.0 & 63.1±8.5\\
Avg. unseen & \textit{IMP} & 65.5±8.5 & 70.4±8.2 & 65.5±10.8 & 59.6±6.2 & 66.1±4.3 & 55.1±10.5\\
\midrule
\textit{FedAvg} & \textit{Fed-MOTIF} & 72.8±5.6 & 78.6±5.9 & \bf 76.7±5.2 & 63.5±6.4 & 72.6±5.8 & \bf 66.0±3.8\\
\textit{FedAvg} & \textit{Fed-IMP} & 71.9±4.2 & 80.3±4.1 & 73.3±8.1 & 58.6±2.0 & 68.8±1.1 & 73.2±2.1\\
\textit{FedSGD} & \textit{Fed-MOTIF} & 39.1±9.8 & 38.6±13.4 & 72.1±3.8 & 52.4±5.6 & 59.9±6.1 & 65.2±6.2\\
\textit{FedSGD} & \textit{Fed-IMP} & 60.6±3.6 & 64.7±5.2 & 66.5±2.7 & 58.7±7.3 & 66.9±7.2 & 48.2±5.0\\
\midrule
All seen & \textit{MOTIF} & \bf 78.5±1.1 & \bf 83.7±1.6 & 72.6±3.7 & \bf 66.2±1.9 & \bf 77.1±2.8 & \bf 65.8±2.8\\
All seen & \textit{IMP} & 64.7±10.4 & 65.4±7.8 & 54.7±6.0 & 58.0±7.4 & 63.3±3.9 & 53.5±5.5\\
\bottomrule
        \end{tabular}
        \caption{
        Results for the \textbf{Predicate Classification} (left) and \textbf{Scene Graph Generation} (right) tasks for Centralized and Federated Learning on the \textbf{PC} dataset). All configurations are run 5 times using random seeds.
        }
        \end{table*}
        
\begin{table*}
        \centering
        \begin{tabular}{ll ccc ccc}
        \toprule
        &&\multicolumn{6}{c}{\textbf{BHSD Dataset}}\\
        \cmidrule(lr){3-8}
        &&\multicolumn{3}{c}{\textbf{Predicate Classification}} & \multicolumn{3}{c}{\textbf{Scene Graph Generation}}\\
        \cmidrule(lr){3-5}\cmidrule(lr){6-8}
        Method & Model & R@8$\uparrow$ & mR@8$\uparrow$ & mAP@8$\uparrow$ & R@8$\uparrow$ & mR@8$\uparrow$ & mAP@8$\uparrow$\\
        \midrule
BHSD dataset & \textit{MOTIF} & 67.9±2.5 & 58.1±4.2 & 36.6±2.3 & 47.8±4.8 & 44.8±6.5 & 26.5±3.0\\
BHSD dataset & \textit{IMP} & 61.7±6.3 & 52.2±6.0 & 34.9±5.9 & 37.2±3.4 & 28.7±5.6 & 21.9±3.8\\
Avg. unseen & \textit{MOTIF} & 51.1±17.7 & 41.0±16.3 & 41.8±10.7 & 47.9±15.8 & 37.4±12.4 & 26.0±9.1\\
Avg. unseen & \textit{IMP} & 49.6±14.9 & 38.5±11.5 & 37.4±8.9 & 41.4±10.6 & 30.9±7.1 & 23.7±9.9\\
\midrule
\textit{FedAvg} & \textit{Fed-MOTIF} & 68.0±5.9 & 54.4±5.9 & 47.8±5.7 & 61.0±4.7 & 47.4±4.3 & 29.0±2.4\\
\textit{FedAvg} & \textit{Fed-IMP} & 69.8±6.9 & 53.9±4.6 & 48.1±5.0 & 49.7±5.8 & 33.3±4.0 & 34.0±5.6\\
\textit{FedSGD} & \textit{Fed-MOTIF} & 32.4±10.2 & 20.9±7.4 & \bf 50.8±17.9 & 39.0±9.9 & 29.1±5.4 & 34.2±6.2\\
\textit{FedSGD} & \textit{Fed-IMP} & 37.4±7.2 & 29.8±3.0 & 34.5±7.4 & 39.0±10.8 & 35.4±10.3 & 27.5±6.2\\
\midrule
All seen & \textit{MOTIF} & \bf 72.4±4.1 & \bf 59.9±5.8 & 42.4±5.3 & \bf 64.1±4.8 & \bf 48.9±5.4 & \bf 29.5±3.6\\
All seen & \textit{IMP} & 53.5±10.1 & 42.2±6.8 & 36.7±6.2 & 45.7±7.9 & 31.4±6.9 & 21.5±3.0\\
\bottomrule
        \end{tabular}
        \caption{
        Results for the \textbf{Predicate Classification} (left) and \textbf{Scene Graph Generation} (right) tasks for Centralized and Federated Learning on the \textbf{BHSD} dataset). All configurations are run 5 times using random seeds.
        }
        \end{table*}
        
\begin{table*}
        \centering
        \begin{tabular}{ll ccc ccc}
        \toprule
        &&\multicolumn{6}{c}{\textbf{CQ500 Dataset}}\\
        \cmidrule(lr){3-8}
        &&\multicolumn{3}{c}{\textbf{Predicate Classification}} & \multicolumn{3}{c}{\textbf{Scene Graph Generation}}\\
        \cmidrule(lr){3-5}\cmidrule(lr){6-8}
        Method & Model & R@8$\uparrow$ & mR@8$\uparrow$ & mAP@8$\uparrow$ & R@8$\uparrow$ & mR@8$\uparrow$ & mAP@8$\uparrow$\\
        \midrule
CQ500 dataset & \textit{MOTIF} & 51.5±7.7 & 55.0±7.3 & 59.0±5.3 & 48.2±8.6 & 51.9±8.2 & 44.2±5.4\\
CQ500 dataset & \textit{IMP} & 53.0±7.5 & 53.9±7.8 & 52.3±7.1 & 36.4±5.1 & 36.9±6.5 & 37.2±11.5\\
Avg. unseen & \textit{MOTIF} & 42.0±14.3 & 45.5±15.9 & 51.0±9.2 & 42.0±12.2 & 46.1±12.9 & 38.4±15.2\\
Avg. unseen & \textit{IMP} & 42.2±12.2 & 44.4±14.0 & 40.6±10.8 & 37.1±9.1 & 39.0±9.1 & 34.7±15.3\\
\midrule
\textit{FedAvg} & \textit{Fed-MOTIF} & 54.4±3.2 & 59.3±3.8 & \bf 60.3±4.0 & 55.5±6.0 & 62.0±5.6 & \bf 53.9±1.8\\
\textit{FedAvg} & \textit{Fed-IMP} & \bf 60.5±1.1 & \bf 66.0±2.1 & 51.4±4.5 & 52.6±2.2 & 55.1±2.3 & 59.4±3.4\\
\textit{FedSGD} & \textit{Fed-MOTIF} & 22.1±8.8 & 21.7±8.6 & 54.0±11.4 & 30.8±3.0 & 33.7±4.5 & 40.2±3.5\\
\textit{FedSGD} & \textit{Fed-IMP} & 32.9±4.9 & 34.3±4.9 & 46.5±8.1 & 34.9±4.3 & 36.6±5.3 & 40.0±2.7\\
\midrule
All seen & \textit{MOTIF} & 59.1±1.9 & 62.8±2.0 & 54.3±7.5 & \bf 59.5±3.0 & \bf 62.9±2.3 & 47.5±3.1\\
All seen & \textit{IMP} & 38.1±7.3 & 39.0±7.4 & 37.6±6.5 & 38.3±2.6 & 38.8±4.2 & 31.4±4.1\\
\bottomrule
        \end{tabular}
        \caption{
        Results for the \textbf{Predicate Classification} (left) and \textbf{Scene Graph Generation} (right) tasks for Centralized and Federated Learning on the \textbf{CQ500} dataset). All configurations are run 5 times using random seeds.
        }
        \end{table*}

\begin{table*}
        \centering
        \begin{tabular}{c ll ccc ccc}
        \toprule
        &&&\multicolumn{6}{c}{\textbf{INST Dataset}}\\
        \cmidrule(lr){4-9}
        \parbox[t]{3mm}{\multirow{2}{*}{\rotatebox[origin=c]{90}{Bias}}}
        &&&\multicolumn{3}{c}{\textbf{Predicate Classification}} & \multicolumn{3}{c}{\textbf{Scene Graph Generation}}\\
        \cmidrule(lr){4-6}\cmidrule(lr){7-9}
        & Method & Model & R@8$\uparrow$ & mR@8$\uparrow$ & mAP@8$\uparrow$ & R@8$\uparrow$ & mR@8$\uparrow$ & mAP@8$\uparrow$\\
        \midrule
        \parbox[t]{3mm}{\multirow{4}{*}{\rotatebox[origin=c]{90}{All datasets}}}
& INST dataset & \textit{MOTIF} & 61.8±5.9 & 65.2±5.5 & 58.1±4.1 & 47.6±6.3 & 49.2±6.2 & 34.1±6.5\\
& INST dataset & \textit{IMP} & 61.9±5.2 & 66.2±5.3 & 49.6±6.3 & 53.1±4.6 & 54.4±5.4 & 22.4±3.2\\
& Avg. unseen & \textit{MOTIF} & 66.2±13.1 & 68.0±11.0 & 59.5±6.9 & 49.9±16.7 & 50.2±15.5 & 38.8±14.5\\
& Avg. unseen & \textit{IMP} & 60.4±13.1 & 62.7±11.8 & 51.8±11.0 & 54.2±16.3 & 55.7±16.1 & 20.0±10.1\\
\midrule
        \midrule
        \parbox[t]{3mm}{\multirow{6}{*}{\rotatebox[origin=c]{90}{Disabled}}}
& \textit{FedAvg} & \textit{Fed-MOTIF} & 76.0±3.5 & \bf76.4±3.3 &\bf 64.8±0.8 & 65.0±3.9 & 63.3±3.7 &\bf 49.6±4.7\\
& \textit{FedAvg} & \textit{Fed-IMP} & 73.8±5.5 & 74.0±4.3 & 62.7±6.3 &\bf 67.6±5.1 & 67.4±5.2 & 20.8±1.4\\
& \textit{FedSGD} & \textit{Fed-MOTIF} & 57.4±14.3 & 59.2±12.6 & 53.4±11.3 & 53.5±5.1 & 52.9±4.7 & 48.5±6.2\\
& \textit{FedSGD} & \textit{Fed-IMP} & 55.0±6.3 & 55.9±5.6 & 55.8±7.3 & 50.1±5.8 & 54.3±4.9 & 10.1±1.5\\
\cmidrule{2-9}
& All seen & \textit{MOTIF} & \bf74.7±4.3 & 75.8±5.4 & 57.9±3.0 & 56.8±6.1 & 58.1±4.9 & 40.3±1.3\\
& All seen & \textit{IMP} & 72.5±5.1 & 75.3±3.5 & 35.6±4.7 & \bf67.4±2.2 & \bf68.4±2.8 & 17.1±1.6\\
\bottomrule
        \end{tabular}
        \caption{
        \textbf{Ablation study}: effect of frequency bias layers (results on the \textbf{INST} dataset). All configurations are run 5 times using random seeds.
        }
        \end{table*}
        
\begin{table*}
        \centering
        \begin{tabular}{c ll ccc ccc}
        \toprule
        &&&\multicolumn{6}{c}{\textbf{PC Dataset}}\\
        \cmidrule(lr){4-9}
        \parbox[t]{3mm}{\multirow{2}{*}{\rotatebox[origin=c]{90}{Bias}}}
        &&&\multicolumn{3}{c}{\textbf{Predicate Classification}} & \multicolumn{3}{c}{\textbf{Scene Graph Generation}}\\
        \cmidrule(lr){4-6}\cmidrule(lr){7-9}
        & Method & Model & R@8$\uparrow$ & mR@8$\uparrow$ & mAP@8$\uparrow$ & R@8$\uparrow$ & mR@8$\uparrow$ & mAP@8$\uparrow$\\
        \midrule
        \parbox[t]{3mm}{\multirow{4}{*}{\rotatebox[origin=c]{90}{All datasets}}}
& PC dataset & \textit{MOTIF} & 72.4±9.1 & 76.9±7.0 & 77.6±5.4 & 56.8±3.6 & 62.5±5.0 & 53.2±7.1\\
& PC dataset & \textit{IMP} & 67.4±3.0 & 68.5±5.4 & 74.3±8.8 & 64.8±7.6 & 67.8±8.7 & 39.6±10.9\\
& Avg. unseen & \textit{MOTIF} & 72.0±7.3 & 76.1±7.3 & 72.0±7.4 & 56.7±8.0 & 63.8±9.2 & 38.9±10.5\\
& Avg. unseen & \textit{IMP} & 65.2±9.6 & 71.1±8.4 & 58.5±13.0 & 55.7±14.7 & 64.1±14.8 & 24.4±7.7\\
\midrule
        \midrule
        \parbox[t]{3mm}{\multirow{6}{*}{\rotatebox[origin=c]{90}{Disabled}}}
& \textit{FedAvg} & \textit{Fed-MOTIF} & 75.6±1.3 & 80.7±2.5 &\bf 74.1±2.9 & 64.1±2.9 & 71.6±3.1 &\bf 50.2±3.1\\
& \textit{FedAvg} & \textit{Fed-IMP} & 65.3±6.1 & 75.4±4.0 & 60.8±5.1 &\bf 73.8±2.3 & 78.9±2.3 & 30.1±2.6\\
& \textit{FedSGD} & \textit{Fed-MOTIF} & 61.4±6.1 & 65.7±6.5 & 68.2±7.3 & 56.6±7.1 & 60.7±5.5 & 44.1±6.3\\
& \textit{FedSGD} & \textit{Fed-IMP} & 57.5±5.4 & 64.1±6.9 & 55.7±5.7 & 35.1±7.0 & 42.4±8.4 & 15.1±7.3\\
\cmidrule{2-9}
& All seen & \textit{MOTIF} &\bf 81.9±3.4 & \bf85.3±2.7 & 70.5±10.7 & 65.8±1.3 & 75.6±0.6 & 34.5±6.2\\
& All seen & \textit{IMP} & 71.4±5.6 & 77.5±3.2 & 37.7±3.0 &\bf 73.9±2.5 &\bf 81.9±2.6 & 22.9±2.4\\
\bottomrule
        \end{tabular}
        \caption{
        \textbf{Ablation study}: effect of frequency bias layers (results on the \textbf{PC} dataset). All configurations are run 5 times using random seeds.
        }
        \end{table*}
        
\begin{table*}
        \centering
        \begin{tabular}{c ll ccc ccc}
        \toprule
        &&&\multicolumn{6}{c}{\textbf{BHSD Dataset}}\\
        \cmidrule(lr){4-9}
        \parbox[t]{3mm}{\multirow{2}{*}{\rotatebox[origin=c]{90}{Bias}}}
        &&&\multicolumn{3}{c}{\textbf{Predicate Classification}} & \multicolumn{3}{c}{\textbf{Scene Graph Generation}}\\
        \cmidrule(lr){4-6}\cmidrule(lr){7-9}
        & Method & Model & R@8$\uparrow$ & mR@8$\uparrow$ & mAP@8$\uparrow$ & R@8$\uparrow$ & mR@8$\uparrow$ & mAP@8$\uparrow$\\
        \midrule
        \parbox[t]{3mm}{\multirow{4}{*}{\rotatebox[origin=c]{90}{All datasets}}}
& BHSD dataset & \textit{MOTIF} & 72.1±4.7 & 59.9±4.6 & 39.1±2.0 & 38.6±4.5 & 36.9±6.5 & 17.5±4.9\\
& BHSD dataset & \textit{IMP} & 64.5±6.2 & 54.6±5.6 & 37.5±4.7 & 42.6±3.3 & 41.4±6.8 & 14.1±2.9\\
& Avg. unseen & \textit{MOTIF} & 54.1±17.7 & 42.6±15.4 & 37.9±6.5 & 32.5±14.4 & 26.4±13.3 & 15.8±10.2\\
& Avg. unseen & \textit{IMP} & 49.7±17.8 & 40.0±14.3 & 35.7±7.1 & 34.1±16.9 & 35.6±18.0 & 8.3±4.2\\
\midrule
        \midrule
        \parbox[t]{3mm}{\multirow{6}{*}{\rotatebox[origin=c]{90}{Disabled}}}
& \textit{FedAvg} & \textit{Fed-MOTIF} & 70.7±4.0 & 55.1±3.6 & \bf44.4±5.4 & 43.6±4.7 & 35.1±5.2 & 18.7±4.2\\
& \textit{FedAvg} & \textit{Fed-IMP} & 64.8±7.6 & 49.7±5.1 & 43.4±2.7 & 51.8±4.1 & 49.7±3.3 & 11.9±2.6\\
& \textit{FedSGD} & \textit{Fed-MOTIF} & 38.9±10.6 & 29.3±9.0 & 36.4±3.8 & 39.6±5.9 & 26.3±5.1 & 14.4±7.5\\
& \textit{FedSGD} & \textit{Fed-IMP} & 34.0±8.3 & 33.4±5.9 & 35.2±4.9 & 30.0±5.8 & 43.7±6.9 & 4.0±1.8\\
\cmidrule{2-9}
& All seen & \textit{MOTIF} & \bf74.6±5.3 & 59.6±7.7 & 39.0±4.3 & 46.7±4.1 & 40.1±3.8 &\bf 23.0±4.0\\
& All seen & \textit{IMP} & 69.9±5.8 &\bf 60.4±5.0 & 31.8±4.2 &\bf 54.5±4.1 & \bf54.5±5.2 & 9.7±1.1\\
\bottomrule
        \end{tabular}
        \caption{
        \textbf{Ablation study}: effect of frequency bias layers (results on the \textbf{BHSD} dataset). All configurations are run 5 times using random seeds.
        }
        \end{table*}
        
\begin{table*}
        \centering
        \begin{tabular}{c ll ccc ccc}
        \toprule
        &&&\multicolumn{6}{c}{\textbf{CQ500 Dataset}}\\
        \cmidrule(lr){4-9}
        \parbox[t]{3mm}{\multirow{2}{*}{\rotatebox[origin=c]{90}{Bias}}}
        &&&\multicolumn{3}{c}{\textbf{Predicate Classification}} & \multicolumn{3}{c}{\textbf{Scene Graph Generation}}\\
        \cmidrule(lr){4-6}\cmidrule(lr){7-9}
        & Method & Model & R@8$\uparrow$ & mR@8$\uparrow$ & mAP@8$\uparrow$ & R@8$\uparrow$ & mR@8$\uparrow$ & mAP@8$\uparrow$\\
        \midrule
        \parbox[t]{3mm}{\multirow{4}{*}{\rotatebox[origin=c]{90}{All datasets}}}
& CQ500 dataset & \textit{MOTIF} & 52.9±7.7 & 56.9±6.9 & 59.3±5.1 & 39.0±3.7 & 42.2±4.0 & 29.8±5.7\\
& CQ500 dataset & \textit{IMP} & 59.9±4.2 & 61.9±5.7 & 51.9±9.8 & 41.1±3.6 & 42.2±3.1 & 21.3±6.8\\
& Avg. unseen & \textit{MOTIF} & 42.5±14.0 & 46.4±14.8 & 51.0±10.9 & 32.0±15.5 & 34.6±16.8 & 23.4±12.0\\
& Avg. unseen & \textit{IMP} & 43.5±13.3 & 46.8±13.7 & 39.6±10.2 & 39.8±19.8 & 41.0±20.1 & 12.6±7.8\\
\midrule
        \midrule
        \parbox[t]{3mm}{\multirow{6}{*}{\rotatebox[origin=c]{90}{Disabled}}}
& \textit{FedAvg} & \textit{Fed-MOTIF} & 56.1±3.7 & 60.7±3.4 &\bf 61.5±3.0 & 52.3±2.1 & 57.2±2.9 &\bf 37.2±2.3\\
& \textit{FedAvg} & \textit{Fed-IMP} & 57.5±5.5 & 60.1±5.9 & 50.9±2.7 &\bf 63.1±4.7 &\bf 65.7±3.6 & 22.5±4.4\\
& \textit{FedSGD} & \textit{Fed-MOTIF} & 29.6±10.6 & 31.6±10.3 & 55.5±12.6 & 32.9±3.5 & 32.2±4.8 & 29.2±9.8\\
& \textit{FedSGD} & \textit{Fed-IMP} & 31.4±4.2 & 35.4±3.9 & 44.0±4.7 & 46.0±7.6 & 46.1±7.3 & 8.3±2.6\\
\cmidrule{2-9}
& All seen & \textit{MOTIF} &\bf 59.0±3.6 & \bf62.5±4.4 & 49.4±9.6 & 46.5±5.2 & 49.7±4.8 & 26.6±2.6\\
& All seen & \textit{IMP} & 57.7±5.2 & 61.4±6.5 & 30.5±5.6 & 57.1±4.3 & 56.2±4.6 & 13.3±1.6\\
\bottomrule
        \end{tabular}
        \caption{
        \textbf{Ablation study}: effect of frequency bias layers (results on the \textbf{CQ500} dataset). All configurations are run 5 times using random seeds.
        }
        \end{table*}

\end{document}